\documentclass[lettersize,journal]{IEEEtran}
\usepackage{amsmath,amsfonts}
\usepackage{algorithmic}
\usepackage{algorithm}
\usepackage{array}
\usepackage[caption=false,font=normalsize,labelfont=sf,textfont=sf]{subfig}
\usepackage{textcomp}
\usepackage{stfloats}
\usepackage{url}
\usepackage{verbatim}
\usepackage{graphicx}
\usepackage{cite}

\usepackage{tabularx}
\usepackage{CJKutf8}
\usepackage{color}
\usepackage{multirow}
\usepackage{stfloats}
\DeclareMathOperator*{\argmax}{arg\,max}
\usepackage{booktabs}
\usepackage{hyperref}
\hypersetup{
hidelinks,
colorlinks=true,
linkcolor=black,
citecolor=black,
urlcolor=black,
}

\usepackage{booktabs} % 绘制更好看的表格
\usepackage{bm} % 给公式加粗

\begin{document}

\title{CPED: A Large-Scale Chinese Personalized and Emotional Dialogue Dataset for Conversational AI}

\author{
Yirong Chen,~\IEEEmembership{Student Member, IEEE}, Weiquan Fan,~\IEEEmembership{Student Member, IEEE},  Xiaofen Xing,~\IEEEmembership{Member, IEEE}, Jianxin Pang, Minlie Huang,~\IEEEmembership{Senior Member, IEEE}, Wenjing Han, Qianfeng Tie, Xiangmin Xu,~\IEEEmembership{Senior Member, IEEE}
        % <-this % stops a space

%\thanks{Manuscript received April 19, 2022. \textit{(Corresponding author: Xiangmin Xu.)}}

%\iffalse % Manuscript received April 19, 2022. 
%\thanks{This work is supported in part by the Key-Area Research and Development Program of Guangdong Province under Grant 2019B010154003, in part by the Science and Technology Project of Guangzhou under Grant 202103010002, in part by the National Natural Science Foundation of China (NSFC) under Grant U1801262, and in part by the Guangdong Provincial Key Laboratory of Human Digital Twin under Grant 2022B1212010004. \textit{(Corresponding author: Xiangmin Xu.)}} % in part by the Fundamental Research Funds for the Central Universities, under Grants 2019PY21,2019MS028, in part by Guangdong Provincial Key Laboratory under grant 2020B121202011,
\thanks{Manuscript updated May 29, 2022. \textit{(Corresponding author: Xiangmin Xu.)}}
\thanks{Y. Chen, W. Fan, X. Xing, W. Han, Q. Tie and X. Xu are with the UBTECH-SCUT Joint Research Lab, Engineering Research Center of Ministry of Education on Human Body Data Perception, School of Electronic and Information Engineering, South China University of Technology, Guangzhou 510641, China (e-mail: eeyirongchen@mail.scut.edu.cn; weiquan.fan96@gmail.com; xfxing@scut.edu.cn; eewenjinghh@mail.scut.edu.cn; 202120112795@mail.scut.edu.cn; xmxu@scut.edu.cn).}

\thanks{J. Pang is with UBTECH Robotics Corp, Shenzhen 518000, China (email:walton@ubtrobot.com).}

\thanks{M. Huang is with CoAI group, State Key Lab of Intelligent Technology and Systems, Beijing National Research Center for Information Science and Technology, Tsinghua University, Beijing 100084, China (e-mail: aihuang@tsinghua.edu.cn).} % DCST, Institute for Artificial Intelligence, 
%\fi
}

% The paper headers
%\markboth{IEEE/ACM Transactions on Audio, Speech and Language Processing,~Vol.~xx, No.~xx, May~2022}
%{CHEN \MakeLowercase{\textit{et al.}}: CPED: A Large-Scale Chinese Personalized and Emotional Dialogue Dataset for Conversational AI}

\IEEEpubid{0000--0000/00\$00.00~\copyright~2022 IEEE}
% Remember, if you use this you must call \IEEEpubidadjcol in the second
% column for its text to clear the IEEEpubid mark.

\maketitle

\begin{abstract}
% 首先说明当前用于对话智能的数据集通常把情感与个性割裂开来
Human language expression is based on the subjective construal of the situation instead of the objective truth conditions, which means that speakers' personalities and emotions after cognitive processing have an important influence on conversation. 
However, most existing datasets for conversational AI ignore human personalities and emotions, or only consider part of them. It's difficult for dialogue systems to understand speakers' personalities and emotions although large-scale pre-training language models have been widely used.
In order to consider both personalities and emotions in the process of conversation generation, we propose CPED, a large-scale Chinese personalized and emotional dialogue dataset, which consists of multi-source knowledge related to empathy and personal characteristic. 
These knowledge covers gender, Big Five personality traits, 13 emotions, 19 dialogue acts and 10 scenes. CPED contains more than 12K dialogues of 392 speakers from 40 TV shows. 
We release the textual dataset with audio features and video features according to the copyright claims, privacy issues, terms of service of video platforms. 
We provide detailed description of the CPED construction process and introduce three tasks for conversational AI, including personality recognition, emotion recognition in conversations as well as personalized and emotional conversation generation. 
Finally, we provide baseline systems for these tasks and consider the function of speakers' personalities and emotions on conversation.
Our motivation is to propose a dataset to be widely adopted by the NLP community as a new open benchmark for conversational AI research. The full dataset is available\footnote{\url{https://github.com/scutcyr/CPED}}.
\end{abstract}

\begin{IEEEkeywords}
Dialogue system, cognitive processing, conversation generation, data collection.
\end{IEEEkeywords}

\section{Introduction}
\IEEEPARstart{O}{pen-domain} conversation systems are of great significance in the application of human-computer interaction, companionship, depression treatment, autism intervention, etc. \cite{zhou2018emotional, zhang2019dialogpt, Zheng2020Persona}.
Driving dialogue systems to learn expression capabilities from a large-scale dialogue corpus, such as OpenSubtitles \cite{J2009News}, Ubuntu Dialogue Corpus \cite{lowe-etal-2015-ubuntu}, STC \cite{shang2015stc}, LCCC \cite{wang2020chinese}, OpenViDial \cite{meng2020openvidial}, etc., is considered to be feasible.

\begin{figure}[htb]
  \centering
  \includegraphics[width=0.48\textwidth]{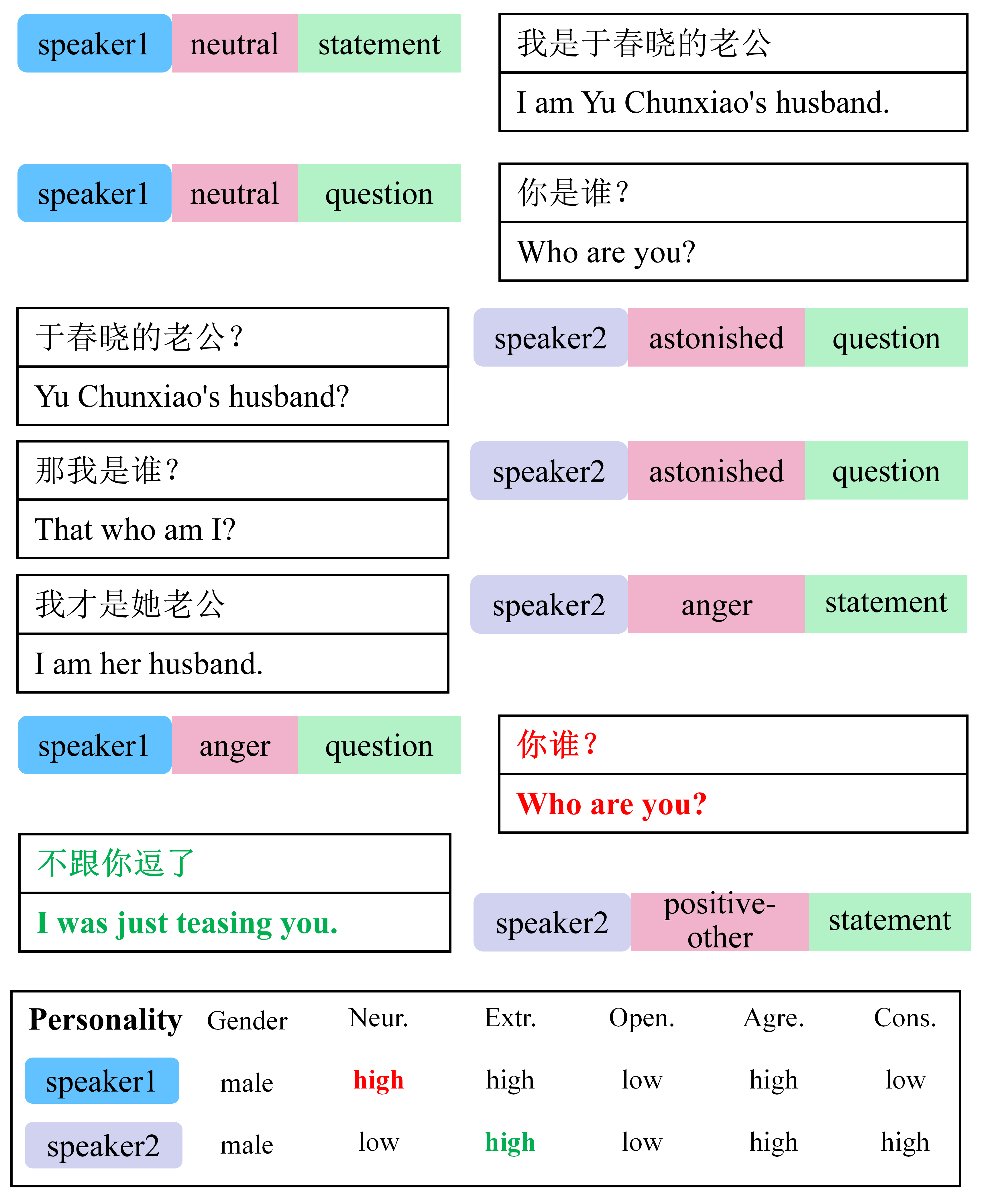}
  \caption{Example from \textbf{CPED} dataset. The dialogue consists of quadruples (speaker, emotion, DA, and utterance) along with speakers' personalities, e.g., gender, Big Five, etc. Note that the emotions or DAs of a speaker would change dynamically during conversation.}
  \label{Dataset_Comparison}
\end{figure}

%% 最新的研究表明人类希望对话系统具备个性化、情感化等拟人化能力。然而没有
\IEEEpubidadjcol
However, if we expect the dialogue systems to possess a good command of personification capabilities, e.g., emotional expression, personality presentation and empathetic conversation, two critical problems need to be tackled: (i) the lack of long-term stable personalities (e.g., gender, age, and Big Five), and (ii) the lack of dynamic emotions or dialogue acts (DAs) during conversation. To the best of our knowledge, dialogue generation models considering emotion and personality as prior knowledge at the same time are currently scarce since no available dialogue dataset simultaneously provides emotional information and personalities of the speakers.

%% 说明在过去的研究中，在对话中把情感或者个性信息作为先验知识引入到对话生成系统中十分重要
%\IEEEpubidadjcol
In a conversation, the participants' expression depends on not only their linguistic context but also the priori personalities and dynamic emotions. For example, in Figure \ref{Dataset_Comparison}, "speaker1" with high \textit{neuroticism} may easily present an angry state in conversation when saying "\begin{CJK}{UTF8}{gbsn}你谁?\end{CJK} (who are you?)". In contrast, "speaker2" with high \textit{extraversion} and low \textit{neuroticism}, may tend to joke during communication, pretending to be Yu Chunxiao's husband to joke with "speaker1". People's personality is imperceptibly affecting their own expression style.
In other words, relying solely on supervised learning on textual contexts is insufficient to model this dialogue generation process. 
Besides, according to the book \textit{Cognitive psychology: Applying the science of the mind}\cite{robinson2016cognitive}, there are also significant differences in Conversation Styles between female and male speakers.
On the whole, only providing large-scale text for training conversation generation models can not make them master human cognitive expression patterns.

%% 引出我们的研究
Therefore, we propose a large-scale \textbf{C}hinese \textbf{P}ersonalized and \textbf{E}motional \textbf{D}ialogue dataset (\textbf{CPED}), which includes the personalities of the speakers, dynamic emotions and DAs of the multimodal dialogue contexts.
CPED, which contains 12K dialogues and 133K utterances, is collected from 40 popular TV series closely related to daily life, making its distribution of personality or emotion close to the real world. We asked the psychology professional annotators to label the emotions and DAs of the speakers through video, audio and text, which is different from DailyDialog \cite{li2017dailydialog} and ESTC \cite{zhou2018emotional}. 
In daily life, speakers may continuously speak in a round of conversation (Figure \ref{Dataset_Comparison}) during which the emotional state or DA state may change several times. Therefore, we divided a turn of dialogue into multiple utterances and annotated emotions and DAs multiple times. Furthermore, we considered gender, age and Big Five personality \cite{big_five} as the basic personality traits.

%% 强调三个特点：多模态、多标注、中文大规模
The contributions of this paper are summarized as follows:
\begin{enumerate}
% 2021.06.28 邢老师建议修改，突出第一点的个性化、情感标注
% 已经修改
\item We build a multiturn \textbf{C}hinese \textbf{P}ersonalized and \textbf{E}motional \textbf{D}ialogue dataset called CPED. To the best of our knowledge, CPED is the first Chinese personalized and emotional dialogue dataset. CPED contains 12K dialogues and 133K utterances with multi-modal context. Therefore, it can be used in both complicated dialogue understanding and human-like conversation generation.

\item CPED has been annotated with 3 character attributes ((name, gender age), Big Five personality traits, 2 types of dynamic emotional information (sentiment and emotion) and DAs. The personality traits and emotions can be used as prior external knowledge for open-domain conversation generation, making the conversation system have a good command of personification capabilities.

% 2021.06.28 邢老师建议修改第3点，重新描述
\item We propose three tasks for CPED: personality recognition in conversations (\textbf{PRC}), emotion recognition in conversations (\textbf{ERC}), and personalized and emotional conversation (\textbf{PEC}). A set of experiments verify the importance of using personalities and emotions as prior external knowledge for conversation generation.
\end{enumerate}

The remainder of this paper is organized as follows: Section II discusses the related work; we then describe the construction process and detailed characteristics of CPED in Section III; definition of personality and emotion recognition in conversations, and corresponding baseline experiments are elaborated in Section IV; definition and baseline experiments of personalized and emotional conversation are described in Section V; applications and limitations of CPED are presented in Section VI; finally, Section VII illustrates the conclusion and future work.

\section{Related Work}
% 参考：https://baijiahao.baidu.com/s?id=1711290199178406354&wfr=spider&for=pc
% 当你阐述一个现象或概念时，使用现在时态
% 当你描述你或别人所做的事情时，使用过去时态
\subsection{Cognitive Psychology Theory for Conversation}

% 基本个性理论
% 16PF、大五、大七、MBTI、九型人格（Enneagram）
\paragraph{Personality Theory}
Allport proposed the \textit{personality traits}~\cite{allport1921personality} in 1921.
%and established a complete personality trait theory~\cite{allport1961pattern} in 1961. 
Allport claims that \textit{personality trait} has the ability to dominate individual behavior, and divides personality traits into two categories: common traits and individual traits. Cattell proposed the Sixteen Personality Factor Questionnaire (16PF) in 1949~\cite{Raymond1995_16pf}.
%after making factor analysis to find 16 independent personality traits~\cite{Raymond1995_16pf}.
Eysenck proposed the structure of personality~\cite{eysenck1953structure} in 1953, and developed the Eysenck Personality Questionnaire (EPQ)~\cite{Eysenck1975Manual} in 1975.
Early lexical studies on personality models~\cite{tupes1992recurrent, norman1963toward} have proved that the terms used to describe personality traits in English are mainly composed of five dimensions, that is named the five factor personality model. McCrae \& Costa (1997) established a five factor personality model based on 16PF factor analysis, which are \textit{Neuroticism}, \textit{Extraversion}, \textit{Openness}, \textit{Agreeableness}, and \textit{Conscientiousness}~\cite{mccrae1997personality}. They also released a NEO Personality Inventory (NEO-PI)~\cite{costa1992four} in 1992 and NEO-PI-R~\cite{mccrae1997personality} in 1997. Typical five factor personality invertories include Hogan Personality Inventory (HPI)~\cite{Hogan1992HPI} and Big Five Inventory (BFI)~\cite{john1990BF}.
Tellegen \& Waller (1987)~\cite{tellegen1987re} used the method of random stratified sampling to select 400 adjectives for self description, and then did factor analysis to obtain the seven dimensions of personality, and put forward the Big Seven factor model of of personality, which are Positive Emotionality (PEM), Nagetive Valence (NVAL), Positive Valence (PVAL), Negative Emotionality (NEM), Dependability (DEP), Agreeableness (AGR), Conventionality (CONV). Among multifarious personality models, the Big Five(BF, also called OCEAN)~\cite{mccrae1997personality, john1990BF} personality model has been proved to have cross-cultural applicability and has been widely used. In addition, Marusic \& Bratko (1998) found that different gender has different distribution in each personality dimension of BF~\cite{marusic1998relations}. Soto et al. (2011) found that BF personality domains has mean-level age differences~\cite{Soto2011AgeDI}. Therefore, gender, age group and BF of the speakers are taken into account in the annotation label.
%In addition, Marusic \& Bratko (1998) found that, femininity shows strong positive relationship with \textit{Agreeableness}, and weak positive relationships with the other four dimensions of BF, while masculinity contributes positively to \textit{Extraversion} and \textit{Conscientiousness}, and negatively to \textit{Neuroticism} and \textit{Agreeableness}~\cite{marusic1998relations}. In other words, different gender has different distribution of each personality dimension.
% 大五人格中文量表：https://mp.weixin.qq.com/s/bQYgB_0WBfyng9LYrjF85A

\paragraph{Emotion Theory}
% https://wenku.baidu.com/view/1556753d13a6f524ccbff121dd36a32d7375c7b0.html
% https://www.ucloud.cn/yun/118846.html
% https://blog.csdn.net/qq_41343328/article/details/122699543
% https://zhuanlan.zhihu.com/p/350243141
% 情绪轮盘：https://www.6seconds.org/2022/03/13/plutchik-wheel-emotions/
More than 90 definitions of ``emotion" have been proposed in the past according to Plutchik's research~\cite{plutchik2001nature}. Izard (1991) divides emotions into three parts: subjective experience, external performance and physiological arousal~\cite{izard1991psychology}. At present, there are two basic representative formats for the classification of emotions: dimensional emotional state (DES) and categorical emotion states (CES)~\cite{9591550}. Russell (1980) defines emotion as two continuous scales: valence and arousal~\cite{russell1980circumplex}. The valence–arousal–dominance space (VAD)~\cite{schlosberg1954three} and the pleasure-arousal-dominance space (PAD)~\cite{mehrabian1996pleasure} are the commonly used DES models, which transform the complex emotions into continuous 3D space. However, it is very difficult to annotate the continuous emotional labels, which consumes a lot of time and human resources, especially for the text. The CES models hold that emotions have completely different structures. Tomkins (1970) believes that there are eight primary emotions~\cite{tomkins1970affects}. In 1971, Ekman \& Friesen proposed the Ekman's six basic emotions, which categorize emotions as: happiness, surprise, anger, disgust, fear and sadness~\cite{ekman1971constants}. In 1980, Robert proposed the Plutchik's Emotion Wheel, which consist 8 basic emotions (anger, disgust, fear, sadness, anticipation, joy, surprise and trust)~\cite{robert1980ii}. Izard (1991) put forward that there are 10 basic emotions~\cite{izard1991psychology}.

\begin{table*}[tb]
\caption{\label{available_dataset}
Comparison among other conversation datasets and CPED. \textit{Modal} denotes the modality of the context (\textit{v}: video, \textit{a}: audio, and \textit{t}: text). \textit{Dial.} denotes the total number of dialogues in the dataset. \textit{Utt.} denotes the total number of utterances in the dataset. \textit{Annotation} indicates how the dataset is labeled in terms of emotion or personality.
}
\centering
\begin{tabular}{m{3.5cm}m{1cm}m{1.1cm}m{1.1cm}m{1.1cm}m{7.5cm}}  % p{65pt}ccccp{200pt}
\toprule
\textbf{Dataset} & \textbf{Lang.} & \textbf{Modal} & \textbf{Dial.} & \textbf{Utt.} & \textbf{Annotation}\\
\midrule
OpenSubtitles~\cite{J2009News} & ML & (\_,\_,t) & - & 11.3M & - \\
Twitter~\cite{sordoni-etal-2015-twitter} & EN & (\_,\_,t) & 4,232 & 33K & - \\
Ubuntu Dialogue Corpus~\cite{lowe-etal-2015-ubuntu} & EN & (\_,\_,t) & 930K & 7.1M & - \\
Cornell Movie Dialogs~\cite{Danescu-Niculescu-Mizil+Lee:11a} & EN & (\_,\_,t) & 220K & 304K & gender and billing-position information of characters \\
OpenViDial~\cite{meng2020openvidial} & EN & (v,{\_},t) & - & 1.1M & - \\
STC~\cite{shang2015stc} & CN  & (\_,\_,t) & 4.4M & 4.6M & - \\
Douban~\cite{wu-etal-2017-douban} & CN & (\_,\_,t) & 1.1M & 6.7M & - \\
LCCC~\cite{wang2020chinese} &  CN & (\_,\_,t) & 12M & 33M & -  \\
WDC-Dialogue~\cite{zhou2021eva} & CN & (\_,\_,t) & 1.4B & 3.0B & - \\
\midrule
IEMOCAP~\cite{Busso2008IEMOCAP} & EN & (v,a,t) & 151 & 7,433 & 10 emotions \\
DailyDialog~\cite{li2017dailydialog} & EN & (\_,\_,t) & 13K &  102K & 7 emotions and 4 DAs and 10 topics \\
Mastodon~\cite{Cerisara2018Mastodon} & EN & (\_,\_,t) & 535 & 2,217 & 3 sentiment tags and 27 DAs  \\
MELD~\cite{poria2019meld} & EN & (v,a,t) & 1,433 & 13,708 & 7 emotions  \\
EmpatheticDialogues~\cite{rashkin-etal-2019-towards} & EN & (\_,\_,t) & 25k & 100K & 32 emotion labels \\
EMOTyDA~\cite{saha2020EMOTyDA} & EN & (v,a,t) & 1,341 & 19,365 & 7 emotions and 12 DAs   \\
ESTC~\cite{zhou2018emotional} &  CN & (\_,\_,t) & 4.4M & 4.5M & 6 emotions (automatically annotated) \\
%PsyQA & CN & (\_,\_,t) & 22k & 78K & 7 mental health support strategy \\
M$^3$ED~\cite{zhao-etal-2022-m3ed} & CN & (v,a,t) & 990 & 24,449 & 7 emotions, role names, ages and genders \\
\midrule
PERSONA-CHAT~\cite{zhang-etal-2018-personalizing} & EN & (\_,\_,t) & 10,981 & 164k & each personas consisting of at least 5 profile sentences \\
PEC~\cite{zhong-etal-2020-towards} & EN & (\_,\_,t) & 355K & 833K & persona sentences for empathetic conversations from subreddits \textit{happy} and \textit{offmychest}\\
MEmoR~\cite{MEmoR} & EN & (v,a,t) & 8,536 & 22,732 & 14 emotions and 3 personality models (16PF, Big Five and MBTI) \\
FriendsPersona~\cite{jiang2020friendspersona} & EN & (v,a,t) & 711 & 8,157 & Big Five personality traits of speaker in a dialogue \\
PELD~\cite{wen-etal-2021-automatically} & EN & (\_,\_,t) & 6,510 & 10,468 & 7 emotions and Big Five personality traits \\
PersonalDialog~\cite{zheng2020personalized} & CN & (\_,\_,t) & 20.83M & 56.25M & 5 personality traits (Age, gender, location, interest, and self descriptions) \\
\midrule
\textbf{CPED(ours)} & \textbf{CN} & \textbf{(v,a,t)} & \textbf{12K} & \textbf{133K} & \textbf{3 sentiments, 13 emotions, 19 DAs, 10 conversation scene, and speaker's personality (Gender, Age, and Big Five)}  \\
\bottomrule
\end{tabular}
\end{table*}

\subsection{Conversation Datasets}
%% 分为open-domain conversation, emotional conversation and personlized conversation.
%It has been demonstrated that large-scale corpus can drive the research of conversation generation system. 
In Table \ref{available_dataset}, we briefly review the available conversation datasets.
\paragraph{Open-domain Conversation Datasets}
There have been various open-domain conversation datasets (Table \ref{available_dataset}(rows 2-10)) over the past few years. These datasets are usually crawled from blogs, forums, or TV series subtitle sites, e.g. OpenSubtitles~\cite{J2009News}, Cornell Movie Dialog Corpus \cite{Danescu-Niculescu-Mizil+Lee:11a}, Ubuntu Dialogue Corpus~\cite{lowe-etal-2015-ubuntu}, Twitter~\cite{sordoni-etal-2015-twitter} and OpenViDial~\cite{meng2020openvidial}. 
%OpenSubtitles~\cite{J2009News} is extracted from the OpenSubtitle website and includes 2.6 billion utterances across 60 languages. The Cornell Movie Dialog Corpus \cite{Danescu-Niculescu-Mizil+Lee:11a} involves 9,035 characters from 617 movies, including 304,713 utterances.
%It also provides the gender and billing-position information of characters, which can be seen as a kind of personality, but almost is not used in dialogue systems. 
%There are also commonly used English textual conversation datasets, e.g., the Ubuntu Dialogue Corpus~\cite{lowe-etal-2015-ubuntu}, Twitter~\cite{sordoni-etal-2015-twitter} and OpenViDial~\cite{meng2020openvidial}. 
%OpenViDial \cite{meng2020openvidial} is collected from English movies and TV series, providing 1.1M visual contexts stored in images and textual dialogues. 
In the field of Chinese conversation generation, the corpus is usually crawled from social media, such as STC \cite{shang2015stc}, the Douban Conversation Corpus \cite{wu-etal-2017-douban}, LCCC \cite{wang2020chinese} and WDC-Dialogue\cite{zhou2021eva}.
Among them, WDC-Dialogue\cite{zhou2021eva} has 1.4 billion dialogues so that the pre-training model can be fully trained in the field of open-domain dialogue generation. 
These datasets do not contain any emotional or personalized annotation information.
Therefore, the dialogue generation model (e.g. DialoGPT \cite{zhang2019dialogpt}, CDialGPT \cite{wang2020chinese}) can only learn personalized or emotional expressions through the dialogue context (single-modal or multi-modal) provided by the corpus.

\paragraph{Emotional Conversation Datasets}
%Recently, the emotional perception and expression ability of the dialogue model have received widespread attention. 
Generally, the emotional perception ability of a dialogue model is defined as the task: emotion recognition in conversations (ERC)~\cite{poria2019meld} or emotion reasoning (ER)~\cite{MEmoR}. Datasets, e.g., IEMOCAP~\cite{Busso2008IEMOCAP}, Mastodon~\cite{Cerisara2018Mastodon}, MELD~\cite{poria2019meld}, EMOTyDA~\cite{saha2020EMOTyDA}, EDA~\cite{bothe2020eda}, MEmoR~\cite{MEmoR} and M$^3$ED~\cite{zhao-etal-2022-m3ed}, are usually used for the ERC or ER task. These datasets generally have small sizes, with fewer than 10K dialogues, making them unsuitable for conversation generation tasks.
Another type of dataset is specifically constructed for emotional conversation generation tasks. For example, DailyDialog~\cite{li2017dailydialog} contains 13K multi-turn dialogues with 102K utterances manually annotated with 7 emotions and 4 DAs. Thus, the dataset is usually used for emotional conversation generation~\cite{Peixiang_et_al_19, HGNN}. EmpatheticDialogues~\cite{rashkin-etal-2019-towards} provides 25K dialogues with 32 types of emotion labels and 2 roles (\textit{speaker} and \textit{listener}) for empathetic conversation. ESTC~\cite{zhou2018emotional}, which is annotated with six emotion categories using the Bi-LSTM emotion classifier based on the STC dataset, is used for Chinese emotional conversation generation.
%PsyQA \cite{psyQA} is a Chinese dataset for mental health support, which is crawled from a Chinese mental health service platform. 
%PsyQA contains 22K questions and 56K answers with 7 mental health support strategy. 
Unfortunately, there is no available large-scale Chinese multi-modal emotional dialogue dataset for emotional conversation generation so far.

\paragraph{Personalized Conversation Datasets}
%% 近年来，人们越来越关注聊天系统的个性化能力。
There are already some datasets related to personalized conversation (in Table~\ref{available_dataset}(rows 19-24)). For example, PERSONA-CHAT~\cite{zhang-etal-2018-personalizing} crowdsourced a set of 1,155 personas and obtained 10,981 dialogs with 164,356 utterances from Turkers assigned a random persona that were asked to chat with others.
In particular, each persona consists of at least 5 profile sentences, just like a small knowledge base that can provide information, such as "I am an artist" or "I like to shi". 
PersonalDialog~\cite{zheng2020personalized}, a Chinese personalized conversation dataset, provides 56.25M utterances from 8.47M speakers who are annotated with personality traits, e.g., age, gender, location, interest tags, etc. 
Specifically, PERSONA-CHAT~\cite{zhang-etal-2018-personalizing} and PersonalDialog~\cite{zheng2020personalized} provide actually character attributes rather than personality traits.
FriendsPersona~\cite{jiang2020friendspersona} is annotated with BF personality traits of speakers, which is used for personality recognition on multiparty dialogues. However, the BF personality traits of speaker in FriendsPersona change in different conversations, which is contradictory to the personality coherence~\cite{roberts2008development}. PELD~\cite{wen-etal-2021-automatically} is proposed for predicting emotion for response using BF personality traits and VAD vector, in which the personality traits are averaged with personality traits of FriendsPersona~\cite{jiang2020friendspersona}. 
MEmoR~\cite{MEmoR}, a recent multimodal emotion reasoning dataset used for the task of multimodal emotion reasoning, provides a multimodal conversation context, 14 fine-grained emotions and 3 types of personalities (16PF, BF and MBTI). MEmoR is mainly used for the task of multimodal emotion reasoning, in which the personalities are used for improving the performance of emotion reasoning. At present, in the field of Chinese conversation, there is a lack of personality related datasets, which hinders the research on personality related tasks, such as personality recognition in conversations.

%4 type of personality knowledge (name, gender, age and Big Five) and 3 type of dynamic emotional information (sentiment, emotion and DA)
With explicit personality and dynamic emotional information, we believe that CPED will provide novel research opportunities and conditions for Chinese open-domain conversation, e.g. personality recognition and emotion recognition on conversations, personalized and emotional conversation.

\section{CPED Dataset}
In this section, we describe the processing stage of constructing the CPED dataset.
To construct a Chinese personalized and emotional dialogue dataset, we collected a large number of TV series related to daily life, and asked the crowdworkers to filter the dialogue segments with abundant emotions and personalities. These dialogue segments were annotated in terms of emotions and personalities by 3 full-time staff of psychology major. In the following, we describe each processing stage of constructing CPED dataset: (1) collecting and prepocessing videos; (2) designing the annotation labels; (3) annotating the emotions and personalities; (4) ensuring annotation quality and re-annotating the overlapping utterance segments.

\subsection{Video Collection and Preprocessing}
\paragraph{Video Source}
In the past, Chinese conversation datasets were obtained by crawling textual dialogues from the Internet. It is difficult to obtain multimodal dialogue data and annotate the emotions and personalities based on multimodal contexts. Therefore, we searched for 100 Chinese TV series closely related to daily life and finally selected 40 TV series that had abundant emotional interaction content and sufficient characters with distinctive personalities.

\paragraph{Dialogue Segment Selection}
We built a Windows application and designed a three-step filtering process to reduce the difficulty of video selection and promote the quality of dialogue segments. Each worker was asked to learn the filtering rules and pass an assessment on which they obtained at least a $98\%$ pass rate in the premarking stage. First, each worker was asked to watch the video and mark the start time and end time of each potential dialogue sample through the developed application. Then, whether every potential dialogue sample was suitable for CPED would be confirmed by another worker. Finally, we split the videos into dialogue segments through the video editing tool \textit{MoviePy}\footnote{\url{https://github.com/Zulko/moviepy}}.

\paragraph{Subtitle Exaction} For most TV series, subtitles are embedded in videos and need to be transcribed to text using the optical character recognition (OCR) technique. We use the video OCR tool \textit{HTWCore}\footnote{\url{https://github.com/xiaopinggai-webrtc/HTWCore}} to generate the subtitles of each dialogue segment. Thus, we obtain the dialogue segments and their subtitles to annotate the emotions, DAs, and personalities.

\begin{figure}[ht]
  \centering
  \includegraphics[width=0.48\textwidth]{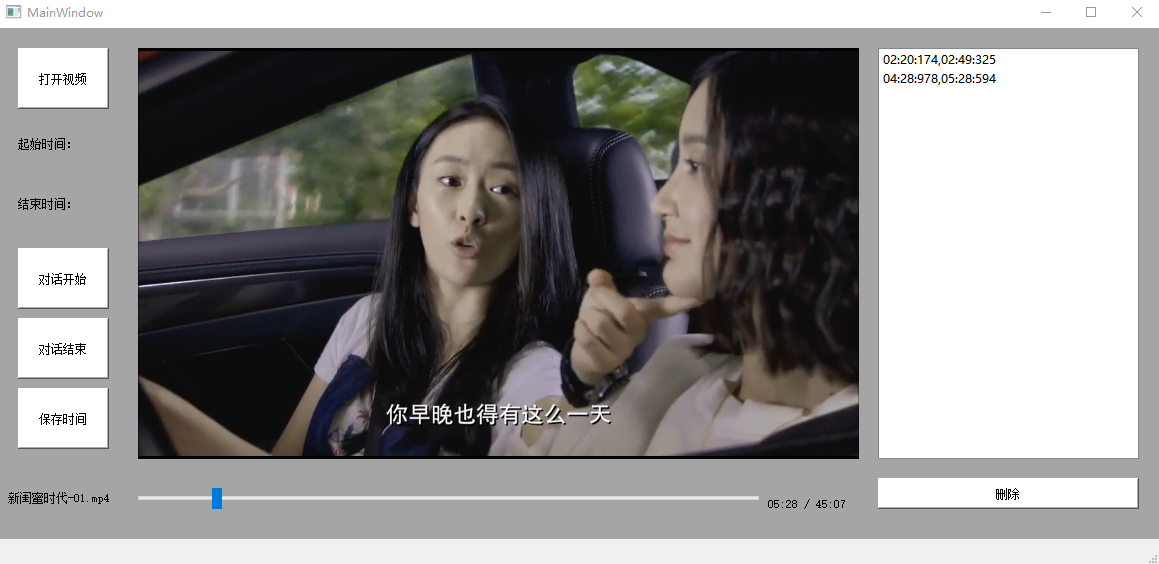}
  \caption{Tools for dialogue segment selection.}
  \label{dialogue_segment}
\end{figure}

\subsection{Annotation Scheme}
\paragraph{Annotation Label}
In order for the dialogue system to learn emotional expression and personalized expression abilities, we provide multiple types of annotation labels listed in Table~\ref{annotation_label_table}: sentiments, emotions, personalities (gender, age group and BF), DAs and scenes. We consider ``\textit{positive, neutral, and negative}'' as the sentiment labels that are the same as MELD~\cite{poria2019meld}. In general, the emotion labels of conversation datasets are considered from among Ekman's six basic emotions (\textit{joy, sad, feared, angry, surprise, and disgusted})~\cite{Ekman1987Universals}. However, the latest studies, e.g., 32 emotion labels in EmpatheticDialogues \cite{rashkin-etal-2019-towards} and 14 emotion labels in MEmoR~\cite{shen2020memor}, show that more fine-grained emotion annotation can contribute to research on emotional reasoning and empathetic conversation. Considering the diversity of emotional tags and the similarity of different tags, we selected 13 emotion labels referring to EmpatheticDialogues~\cite{rashkin-etal-2019-towards} and 19 DA labels referring to the SWBD-DAMSL tag-set \cite{Jurafsky1997SWBD} based on the characteristics of Chinese open-domain conversation. In particular, we have added two special labels, ``\textit{other-positive}'' and ``\textit{other-negative}'', which allow uncommon emotions to be included. Personality is complex and changeable, and there is no unified trait set of personality.
Different from PERSONA-CHAT~\cite{zhang-etal-2018-personalizing} and PersonalDialog~\cite{zheng2020personalized}, we consider gender, age and BF personality as the basic personality traits. According to the \textit{Developmental Psychology}, the age groups are divided into: children ($<11$), teenager ($12-20$), young ($21-39$), middle-aged ($40-60$) and elderly ($>60$). Following Dailydialog~\cite{li2017dailydialog}, we label each dialogue as one of ten dialogue scene categories.

\begin{table}[tb]
\caption{\label{annotation_label_table} Annotation labels of the proposed dataset.}
\centering
\begin{tabular}{lm{5.5cm}c}
\toprule 
\textbf{\# of annos.} & \textbf{Labels} & \textbf{Num.} \\ \midrule
Sentiment & \textit{positive, neutral, and negative} & 3 \\
Emotion & \textit{happy, grateful, relaxed, other-positive, neutral, angry, sad, feared, depressed, disgusted, astonished, worried and other-negative} & 13 \\
\midrule
Gender & \textit{male, female, and unknown} & 3 \\
Age group & \textit{children, teenager, young, middle-aged, elderly and unknown} & 6 \\
Big Five & \textit{high, low, and unknown} & 3 \\
\midrule
DA & \textit{greeting (g), question (q), answer (ans), statement-opinion (sv), statement-non-opinion (sd), apology (fa), command (c), agreement/acceptance (aa), disagreement (dag), acknowledge (a), appreciation (ba), interjection (ij), conventional-closing (fc), thanking (ft), quotation ( ${\hat{}}$q), reject(rj), irony (ir), comfort (cf) and other (oth)} & 19 \\
\hline
Scene & \textit{home, office, school, mall, hospital, restaurant, sports-venue, entertainment-venue, car, outdoor and other-scene} & 11 \\

\bottomrule
\end{tabular}
\end{table}

\paragraph{Annotation Process}
% 2021.06.26 修改至这里
The annotation process is divided into two stages: (1) utterance-level annotation and (2) speaker-level annotation. First, we ask annotators to label the sentiments, emotions, DAs and scenes of each utterance. Second, when the dialogue samples of a TV series have been annotated, the experts are asked to annotate the \textit{gender}, \textit{age group} and \textit{Big Five} of each character that appears in the dialogue samples. In particular, the \textit{Chinese Big Five Inventory-2 (Chinese BFI-2)}\cite{zhang2021big} proposed by Zhang \textit{et al.} is used for calculating the scores of Big Five personalities. Annotators were asked to fill \textit{Chinese BFI-2} for each speaker. The normalized average of the final score is used to judge the personality traits (\textit{high}, \textit{unknown} or \textit{low}).

\subsection{Annotation Tool}
\begin{figure}[htbp]
  \centering
  \includegraphics[width=0.48\textwidth]{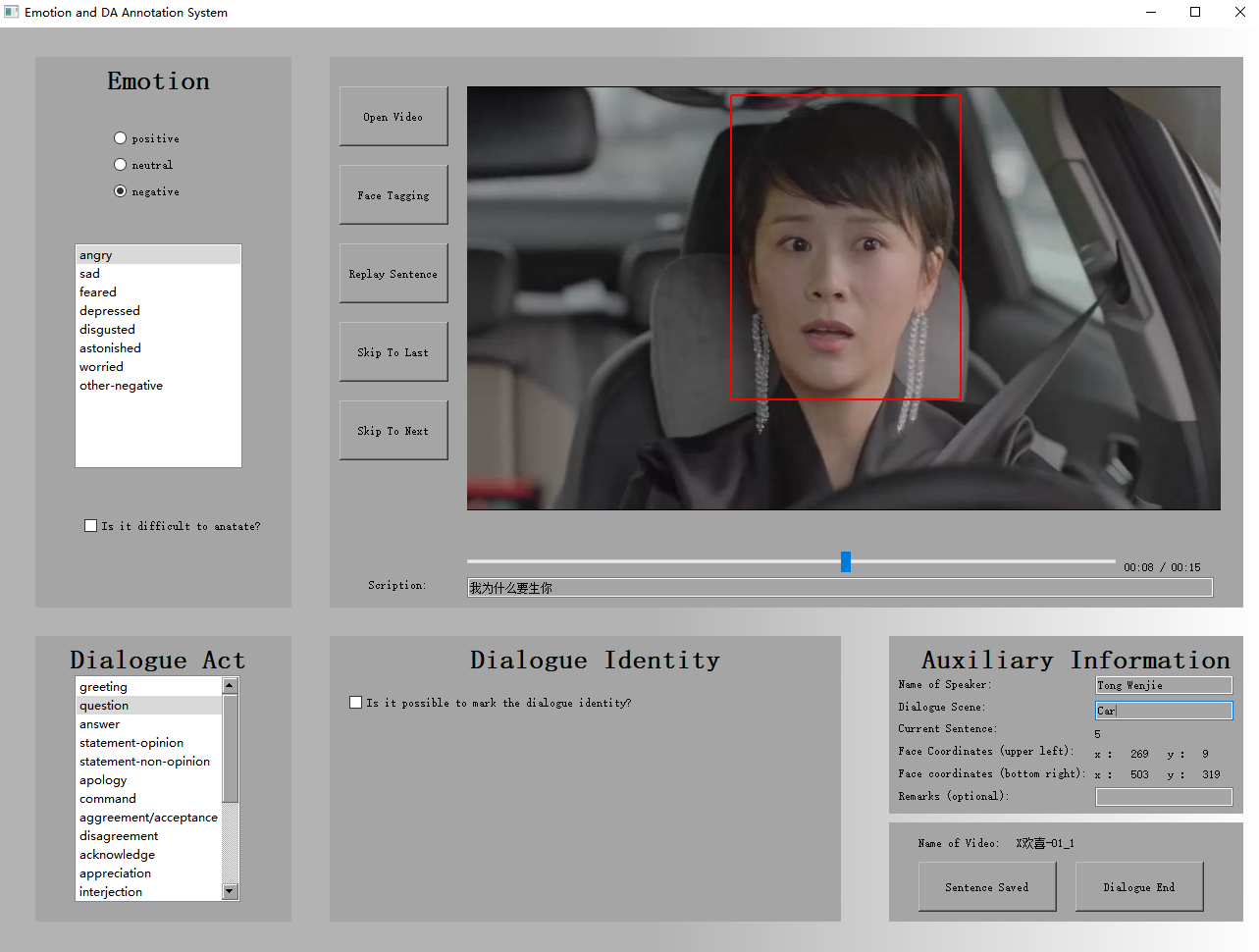}
  \caption{Conversation annotation application.}
  \label{conversation_annotation}
\end{figure}

We built two Windows applications for dialogue segment and annotation by using the \textit{PyQt}\footnote{\url{https://www.riverbankcomputing.com/software/pyqt}} tool, as shown in Figure \ref{dialogue_segment} and Figure \ref{conversation_annotation}. In the dialogue segment cutting stage, the annotators click the button "\begin{CJK}{UTF8}{gbsn}\small{打开视频}\end{CJK} (open video)", select an original video (about 40min), and then mark the start time and end time of the dialogue segment by repeatedly clicking the buttons "\begin{CJK}{UTF8}{gbsn}\small{对话开始}\end{CJK} (start of dialogue)" and "\begin{CJK}{UTF8}{gbsn}\small{
对话结束}\end{CJK} (end of dialogue)".

As shown in Figure \ref{conversation_annotation}, annotators click "open video" to open a short dialogue video and the corresponding subtitle file. For each sentence, annotators need to select the sentiment, emotion and dialogue act. Meanwhile, they need to fill in the speaker's name of each sentence and the scene of the whole dialogue sample.

\subsection{Annotation Quality Control}
To guarantee quality, we recruit three psychology experts who have a wealth of prior knowledge and experience for discriminating emotion, DA and personality. We jointly formulated labeling rules and labeling examples and randomly selected 200 samples for 3 rounds of prelabeling, thereby reducing the discrepancy in labeling by discussing and improving the annotation scheme. Following \cite{poria2019meld}, experts are required to annotate utterances with multi-modal information that combines video, facial expressions, audio and text, which can help improve the emotional annotation accuracy. Each utterance was annotated by 3 experts, and the majority rule was used to determine the final labels. If the labeling results of the three experts are inconsistent, they needed to reannotate those utterances to find a ``common'' annotation. Finally, samples that still could not be labeled uniformly were discarded. In addition, since some speakers rarely speak, they will be uniformly defined as ``\begin{CJK}{UTF8}{gbsn}其他\end{CJK} (other)'', of which the gender, age group, and Big Five personality will be annotated as ``unknown''. Finally, we include a total of 11,835 dialogues with multi-source knowledge.

\begin{table}[ht]
\caption{\label{dialogue_overlap_processing} Example of utterance overlap that need to be cut into multiple utterances correctly.}
\centering
\begin{tabular}{m{5.2cm}c}
\toprule 
\textbf{Utterance} & \textbf{Speaker}  \\ 
\midrule
\begin{CJK}{UTF8}{gbsn}\textcolor[RGB]{0,0,255}{多大的事}\textcolor[RGB]{255,0,0}{你知道的我把握不好尺度}\end{CJK} & \begin{CJK}{UTF8}{gbsn}胡一菲\end{CJK} \\

\textcolor[RGB]{0,0,255}{Big deal.}\textcolor[RGB]{255,0,0}{You know, I can't hold the scale.} & Hu Yifei \\

\midrule
\begin{CJK}{UTF8}{gbsn}多大的事啊\end{CJK} & \begin{CJK}{UTF8}{gbsn}胡一菲\end{CJK} \\
Big deal. & Hu Yifei \\
\begin{CJK}{UTF8}{gbsn}你知道的我把握不好尺度\end{CJK} & \begin{CJK}{UTF8}{gbsn}陆展博\end{CJK} \\
You know, I can't hold the scale. & Lu Zhanbo \\
\bottomrule
\end{tabular}
\end{table}

\paragraph{Utterance Overlap Processing}
% 2021.06.27改到这里
Automatic subtitle extraction will be accompanied by \textit{utterance overlap}, which means that one utterance contains the content of two speakers talking (Table \ref{dialogue_overlap_processing}). The statistics indicated that there were 4,613 \textit{utterance overlaps} identified by annotators during the construction of the entire dataset. These utterance samples were correctly cut into multiple utterances, and the emotions and DAs were respectively reannotated.

\begin{figure*}[ht]
  \centering
  \includegraphics[width=\textwidth]{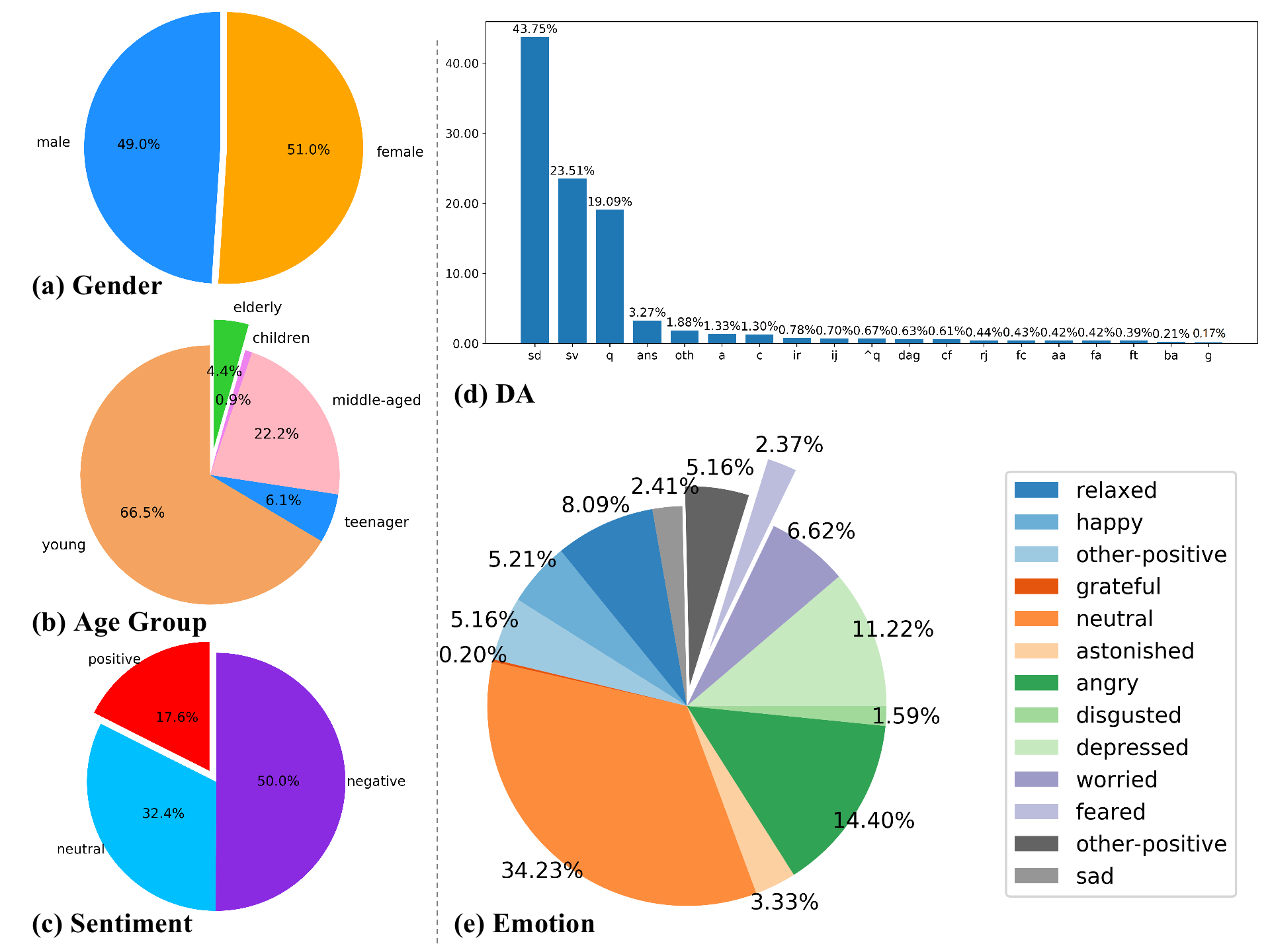}
  \caption{Distribution of Gender, Age Group, Sentiment, Emotion and DA in CPED Dataset.}
  \label{Statistics}
\end{figure*}

\subsection{Corpus Exploration}
\paragraph{Dataset Split}
We randomly split the CPED dataset into three sets: train, valid and test according to the ratio of 7:1:2. In order to avoid data leakage, the split of the dataset is based on TV series, which ensures that the speakers in the training set will not appear in the valid/test set.

\begin{table}[htbp]
\caption{\label{dataset_statistics} Summary of CPED dataset statistics. \textit{utt.}, \textit{dial.}, \textit{emot.} refer to utterance, dialogue, emotion. (v,a,t)=(visual, audio, text).}
\centering
\begin{tabular}{lccc}
\toprule
\textbf{Statistics} & \textbf{Train} & \textbf{Valid} & \textbf{Test} \\ 
\midrule
\# of modalities & (v,a,t) & (v,a,t) & (v,a,t) \\
\# of TV plays & 26 & 5 & 9 \\
\# of dialogues & 8,086 & 934 & 2,815 \\
\# of utterances & 94,187 & 11,137 & 27,438 \\
\# of speakers & 273 & 38 & 81 \\
Avg. \# utt. per dial. & 11.6 & 11.9 & 9.7 \\
Max \# utt. per dial. & 75 & 31 & 34 \\
%Avg./Max \# utt. per dial. & 11.6/75 & 11.9/31 & 9.7/34 \\
Avg. emot. per dial. & 2.8 & 3.4 & 3.2 \\
Avg. DAs per dial. & 3.6 & 3.7 & 3.2 \\
Avg. utt. length & 8.3 & 8.2 & 8.3 \\
Max utt. length & 127 & 42 & 45 \\
%Avg./Max utt. length & 8.3/127 & 8.2/42 & 8.3/45 \\
Avg. duration & 2.1s & 2.12s & 2.21s \\
\bottomrule
\end{tabular}
\end{table}

\begin{figure}[ht]
  \centering
  \includegraphics[width=0.48\textwidth]{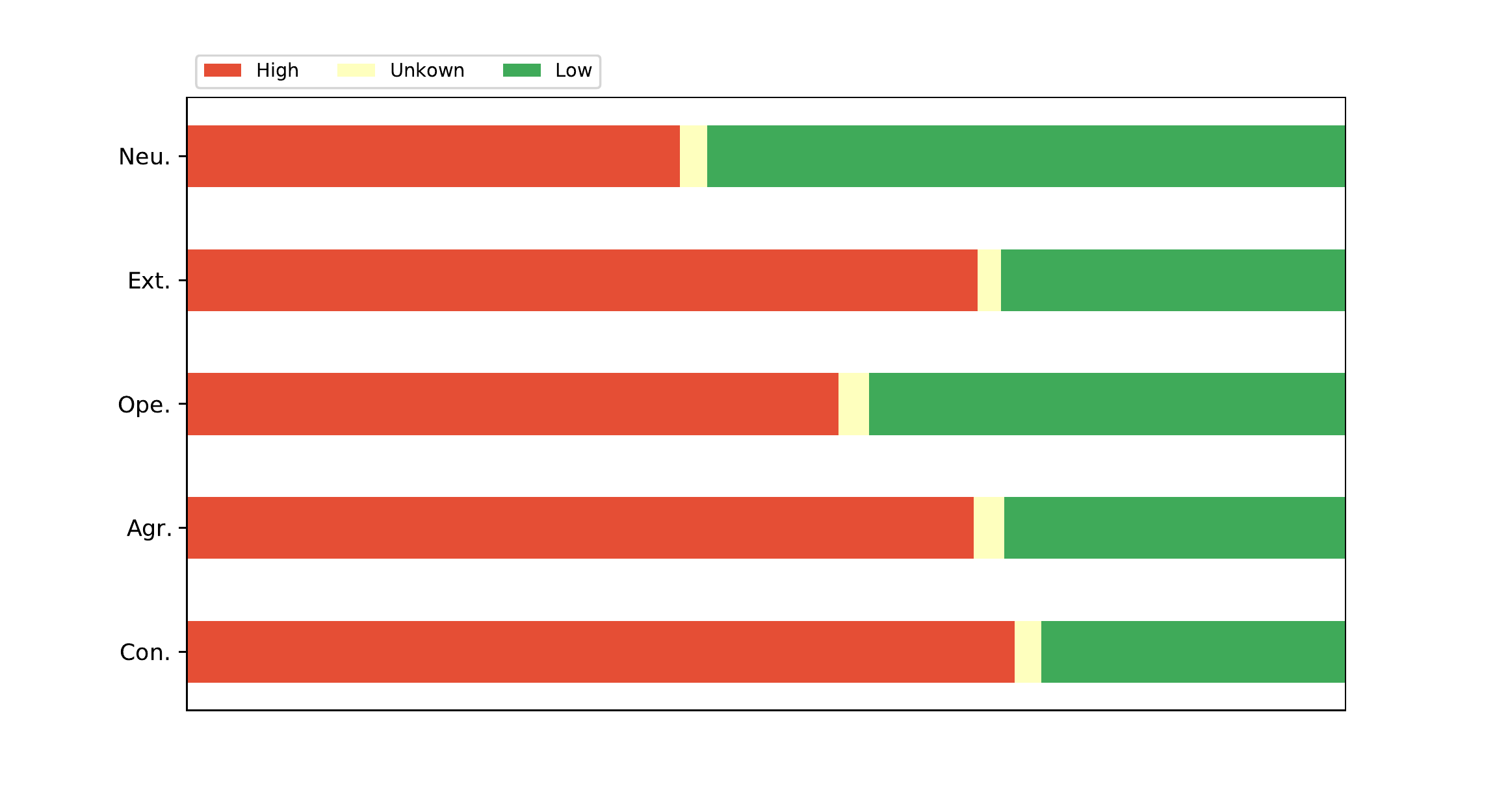}  \caption{Distribution statistics of Big Five (Neu.: Neuroticism, Ext.: Extraversion, Ope.: Openness, Agr.: Agreeableness, and Con.: Conscientiousness).}
  \label{bigfive_statistics}
\end{figure}

\paragraph{Dataset Statistics}
Figure \ref{Statistics} presents the distribution of the genders, ages groups, sentiments, emotions and DAs of the CPED dataset. The ratio of males to females is close to 1:1, which makes the distribution of personality and emotion close to the real world. Similar to other conversation datasets, the distribution of emotion and DA labels are unbalanced. Among them, ``neutral'' accounts for 32.4\% of all emotions. The statistics of CPED are listed in Table \ref{dataset_statistics}. The average numbers of emotions per dialogue, i.e., the number of different emotion categories, are 2.8, 3.4 and 3.2 in training/validation/testing samples. The average DAs per dialogue are 3.6, 3.7, and 3.2 in training/validation/testing samples. As shown in Figure \ref{bigfive_statistics}, the proportion of \textit{high} is higher than that of \textit{low} in Extraversion, Openness, Agreeableness and Conscientiousness, while lower in Neuroticism.

\paragraph{Dataset sample}
Each sample in the CPED dataset is composed of a series of utterance-level videos, textual context and multiple annotation results (name, gender, age group, Big Five personality, sentiment, emotion and DA). Table \ref{dataset_format} shows the final format of one utterance on the CPED dataset in which researchers can obtain the audio file and video file corresponding to the utterance through \textit{Utterance\_ID}.

\begin{table}[htbp]
\caption{\label{dataset_format}
CPED dataset format for an utterance. Big Five = (neuroticism, extraversion, openness, agreeableness, and conscientiousness)}
\centering
\begin{tabular}{ll}
\toprule
\textbf{One utterance} &  \\
\midrule
\textbf{Dialogue\_ID} & 01\_000 \\
\textbf{Utterance\_ID} & 01\_000\_000 \\
\textbf{Speaker} & \begin{CJK}{UTF8}{gbsn}童文洁\end{CJK}(Tong Wenjie) \\
\textbf{Gender} & female \\
\textbf{Age} & middle-aged \\
\textbf{Sentiment} & neutral \\
\textbf{Emotion} & neutral \\
\textbf{Big Five} & (high, high, low, low, high) \\
\textbf{DA} & greeting \\
\textbf{Scene} & other-venue \\
\textbf{Utterance} & \begin{CJK}{UTF8}{gbsn}真巧\end{CJK}(What a coincidence) \\
\bottomrule
\end{tabular}
\end{table}

\section{Personality and Emotion Recognition in Conversations}
We are committed to making the dialogue system acquire cognitive ability like human, including understanding the personalities of the speaker and speaker's current emotion through conversations. Therefore, we research two subtasks respectively: personality recognition in conversations (PRC) and emotion recognition in conversations (ERC).

% 个性计算与情感计算
\subsection{Personality Recognition in Conversations (PRC)}
%\subsection{Task Definition}

\begin{figure}[htbp]
  \centering
  \includegraphics[width=0.45\textwidth]{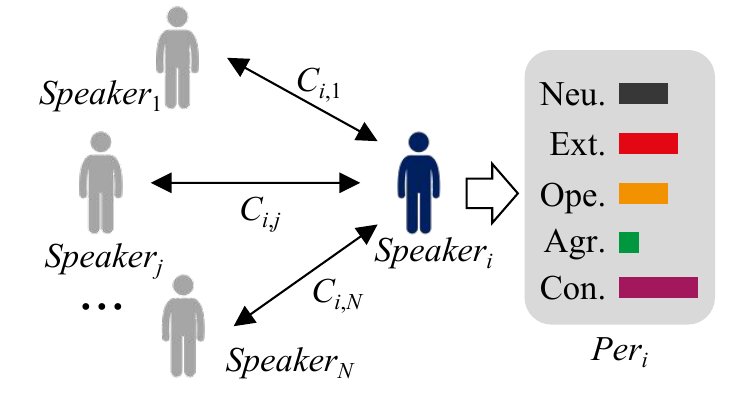}
  \caption{Personality recognition in $1-N$ conversations.}
  \label{personality_recognition_task}
\end{figure}

\paragraph{Task Definition}
Given a speaker's conversation with others, it is required to recognize the speaker's personality traits through the conversation record, which includes two scenarios, (1) $1-1$ conversations: the robot recognizes the personality traits of the speaker through the conversation between them (e.g., psychological counseling), (2) $1-N$ conversations (see Figure \ref{personality_recognition_task}): the robot listens to the speaker's conversations with other $N$ people and then recognizes the speaker's personality traits (e.g., group chatbot, home service robot). Since $1-N$ includes the case of $1-1$, we only discusses PRC in $1-N$ conversations. The task of PRC in $1-N$ conversations can be formulated as:
\begin{equation}
Per_i = \argmax_{Per'_i}P(Per'_i|C_{i,j},\cdots,C_{i,N}), \label{prc_definition}
\end{equation}
where $Per_i=[Neu, Ext, Ope, Agr, Con]$ is a 5-dimensional vector representing Neuroticism, Extraversion, Openness, Agreeableness, and Conscientiousness. $C_{i,j}$ is the conversations between $Speaker_i$ and $Speaker_j$ ($1 \leq j \leq N$).

\paragraph{Baseline Models}
Several benchmack models are provided for PRC task, including 

1) \textbf{BERT\bm{$^{s}$}} only uses the concatenation of the utterances of the target $Speaker_i$~\cite{jiang2020friendspersona} as the input to BERT~\cite{devlin2019bert}, while \textbf{BERT\bm{$^{c}$}} uses the full conversation $C_{i,j}$ in the natural order. The personality $Per_i$ of $Speaker_i$ is calculated as follows:
\begin{align}
h^{[CLS]}_{i,j} &= BERT(T_{i,j}), \\
h^{[CLS]} &= [h^{[CLS]}_{i,1},\cdots,h^{[CLS]}_{i,N}], \\
Per_{i,k} &= MLP_{k}(Avgpooling(h^{[CLS]})),
\end{align}

where $T_{i,j}$ is $U_{i,j}$ for \textbf{BERT\bm{$^{s}$}} and $C_{i,j}$ for \textbf{BERT\bm{$^{c}$}}. $U_{i,j}=[[CLS],u_1,u_2,\cdots,u_m]$ is the concatenation of the utterances of the target $Speaker_i$~ in conversation $C_{i,j}$. $[CLS]$ is the special token of BERT. $MLP_{k}$ is a multi-layer perceptron. $Avgpooling$ means average pooling. $k$ is the index of personality in different dimensions.

2) \textbf{BERT\bm{$_{ssenet}^{c}$}} uses the full conversation text as the input to BERT and a shared SeNet~\cite{Hu_2018_CVPR} as the feature fusion layer of local personality features of different conversations, which can be formulated as:
\begin{align}
% h^{[CLS]}_{i,j} &= BERT(C_{i,j}), \\
% h^{[CLS]} &= [h^{[CLS]}_{i,1},\cdots,h^{[CLS]}_{i,N}], \\
Per_{i,k} &= MLP_{k}(SeNet(h^{[CLS]})),
\end{align}

where $k \in [1,5]$ is the dimension index of personalities.

3) \textbf{BERT\bm{$_{senet}^{c}$}} uses the full conversation text as the input to BERT and five independent SeNets as the feature fusion layer of local personality features of different conversations, which can be formulated as:
\begin{align}
% h^{[CLS]}_{i,j} &= BERT(C_{i,j}), \\
% h^{[CLS]} &= [h^{[CLS]}_{i,1},\cdots,h^{[CLS]}_{i,N}], \\
Per_{i,k} &= MLP_{k}(SeNet_{k}(h^{[CLS]})),
\end{align} 

\begin{table}[htbp]
\caption{\label{personality_recognition} Personality Recognition in CPED (Neu.: Neuroticism, Ext.: Extraversion, Ope.: Openness, Agr.: Agreeableness, Con.: Conscientiousness and Avg.: Average).}
\centering
\begin{tabular}{m{1.25cm}m{0.48cm}m{0.48cm}m{0.48cm}m{0.48cm}m{0.48cm}m{0.48cm}m{1.1cm}}
\toprule
\multirow{2}{*}{\textbf{Model}} & \multicolumn{6}{c}{\textbf{Accuracy}} & \multirow{2}{*}{\textbf{Macro-F1}}
     \\\cmidrule(lr){2-7}     & \textbf{Neu.} & \textbf{Ext.} & \textbf{Ope.} & \textbf{Agr.} & \textbf{Con.} & \textbf{Avg.}\\\midrule
BERT$^{s}$ & 50.75 & 78.08 & \textbf{57.93} & 85.76 & \textbf{63.60} & 67.23 & 72.93 \\
BERT$^{c}$ & \textbf{55.29} & 78.08 & 53.90 & 80.98 & 63.35 & 66.32 & 72.69 \\
BERT$_{senet}^{c}$ & 53.40 & 77.71 & 55.42 & 81.99 & 61.59 & 66.02 & 71.89 \\
BERT$_{ssenet}^{c}$ & 53.27 & \textbf{78.21} & 55.42 & \textbf{85.89} & 63.48 & \textbf{67.25} & \textbf{74.08} \\ % 五个维度共享一个senet
\bottomrule
\end{tabular}
\end{table}

\paragraph{Experimental Results}
Table \ref{personality_recognition} shows the performance of the baseline models on PRC task. Comparing the performance of BERT$^{s}$ and BERT$^{c}$, it can be found that BERT$^{s}$ achieves the better performance in Openness (57.93\%) and Agreeableness (85.76\%) while BERT$^{c}$ achieves the better performance in Neuroticism (55.29\%). In the two dimensions of Extraversion and Conscientiousness, the performance of BERT$^{s}$ and BERT$^{c}$ is basically the same. Future work should needs to set up different recognition frameworks for different dimensions of BF, which requires further psychological analysis. Among the four baseline models, BERT$_{ssenet}^{c}$ achieves the best performance in Macro-F1 and average accuracy. Among all the dimensions, the performances on Extraversion and Agreeableness are much higher than others because the two dimensions are more consistent when the speaker communicates with different characters. The future research on PRC in $1-N$ conversations needs to further consider the demonstrated personalities are different when the speaker talk with different characters.

\subsection{Emotion Recognition in Conversations (ERC)}
\paragraph{Task Definition} ERC task focuses on identifying the sentiment-level or emotion-level labels $e_M$ of the utterance $u_{M}$ according to the conversation context $U={u_1,\cdots,u_M}$:

\begin{equation}
e_M = \argmax_{e'}P(e'|u_{1},\cdots,u_{M-1},u_{M}), \label{erc_definition}
\end{equation}

where $u_{i}$ is the utterance contains several tokens spoken by $Speaker(u_{i})$. $M$ is the number of the utterances in the conversation.

\paragraph{Baseline Models}
In order to provide an effective benchmark, we consider two types of benchmark models: ERC models with current single utterance as input and ERC models with current utterance and dialogue history as input.

1) ERC models with current single utterance $u_{M}$ as input, include \textbf{TextCNN}~\cite{TextCNN}, \textbf{TextRNN}~\cite{TextRNN}, \textbf{TextRCNN}~\cite{TextRCNN}, \textbf{FastText}~\cite{joulin-etal-2017-bag}, \textbf{BERT}~\cite{devlin2019bert}. They use sentence-level language model $LM$ to obtain the representation of $u_{M}$, and then use $MLP$ to predict the emotion $e_M$ as follows:
\begin{align}
h &= LM(u_{M}), \\
e_M &= MLP(h)),
\end{align}

2) ERC models with both current utterance and dialogue history include \textbf{bcLSTM}~\cite{poria-etal-2017-context}, \textbf{DialogueRNN}~\cite{majumder2019dialoguernn}, \textbf{DialogueGCN}~\cite{ghosal2019dialoguegcn}, \textbf{DialogXL}~\cite{shen2021dialogxl} and \textbf{EmoBERTa}~\cite{taewoon2021emoberta}. \textbf{bcLSTM} is a bi-directional contextual LSTM model that has two unidirectional and opposite-direction LSTMs stacked together. \textbf{DialogueRNN} is a classic and efficient algorithm for ERC, which uses three gated recurrent units (GRU) to model the dialogue process, including the global GRU, the party GRU and the emotion GRU. \textbf{DialogueGCN} captures richer contextual information by considering the speaker information of the utterance and the relative positions of the target utterance and the context, which has three stages, consisting of sequential context encoding, speaker-level context encoding and classification. \textbf{DialogXL} used pre-trained language models for ERC, in which the memory-saving utterance recurrence mechanism and dialog-aware self-attention are used. \textbf{EmoBERTa} is a speaker-aware model based on RoBERTa, which prepends speaker names to utterances. We also proposed a baseline model \textbf{BERT+AVG+MLP} based on BERT~\cite{devlin2019bert}, which adds all the speaker names to the special token dictionary and splices the speaker names and utterances sequentially as the input of BERT. The average pooling of hidden-layer output of BERT then inputs to multi-layer perceptron (MLP) to predict the emotion labels.

\begin{table}[htbp]
\caption{\label{sentiment_level_recognition} Sentiment Recognition in CPED (Neg.: Negative, Neu.: Neutral, Pos.: Positive and Avg.: Average).}
\centering
\begin{tabular}{lccccc}
\toprule
\multirow{2}{*}{\textbf{Model}} & \multicolumn{4}{c}{\textbf{Accuracy}} & \multirow{2}{*}{\textbf{Macro-F1}}
     \\\cmidrule(lr){2-5}     & \textbf{Neg.} & \textbf{Neu.} & \textbf{Pos.} & \textbf{Avg.}\\\midrule
TextCNN~\cite{TextCNN} & 64.51 & 24.56 & 14.04 & 48.90 & 34.37\\
TextRNN~\cite{TextRNN} & 62.69 & 33.21 & 15.32 & 47.89 & 37.07\\
TextRCNN~\cite{TextRCNN} & 64.04 & 31.03 & 18.78 & 49.13 & 37.95\\
FastText~\cite{joulin-etal-2017-bag} & \textbf{65.30} & 24.76 & 0.95 & 48.62 & 30.33\\
BERT~\cite{devlin2019bert} & 59.97 & \textbf{42.98} & 32.60 & 48.96 & 45.18\\
\midrule
bcLSTM~\cite{poria-etal-2017-context} & 59.06 & 38.86 & 28.28 & 49.65 & 45.40 \\
DialogueRNN~\cite{majumder2019dialoguernn} & 59.06 & 36.16 & \textbf{44.97} & 48.57 & 44.11 \\
DialogueGCN~\cite{ghosal2019dialoguegcn} & 60.65 & 36.99 & 40.19 & 47.69 & 45.12 \\
EmoBERTa~\cite{taewoon2021emoberta} & 59.24 & 35.02 & 41.53 & 48.09 & 44.60 \\
DialogXL~\cite{shen2021dialogxl} & 60.45 & 41.45 & 41.74 & 51.24 & 46.96 \\
BERT+AVG+MLP & 61.40 & 40.10 & 42.95 & \textbf{51.50} & \textbf{48.02} \\
\bottomrule
\end{tabular}
\end{table}

\begin{figure}[htbp]
  \centering  
  \includegraphics[width=0.45\textwidth]{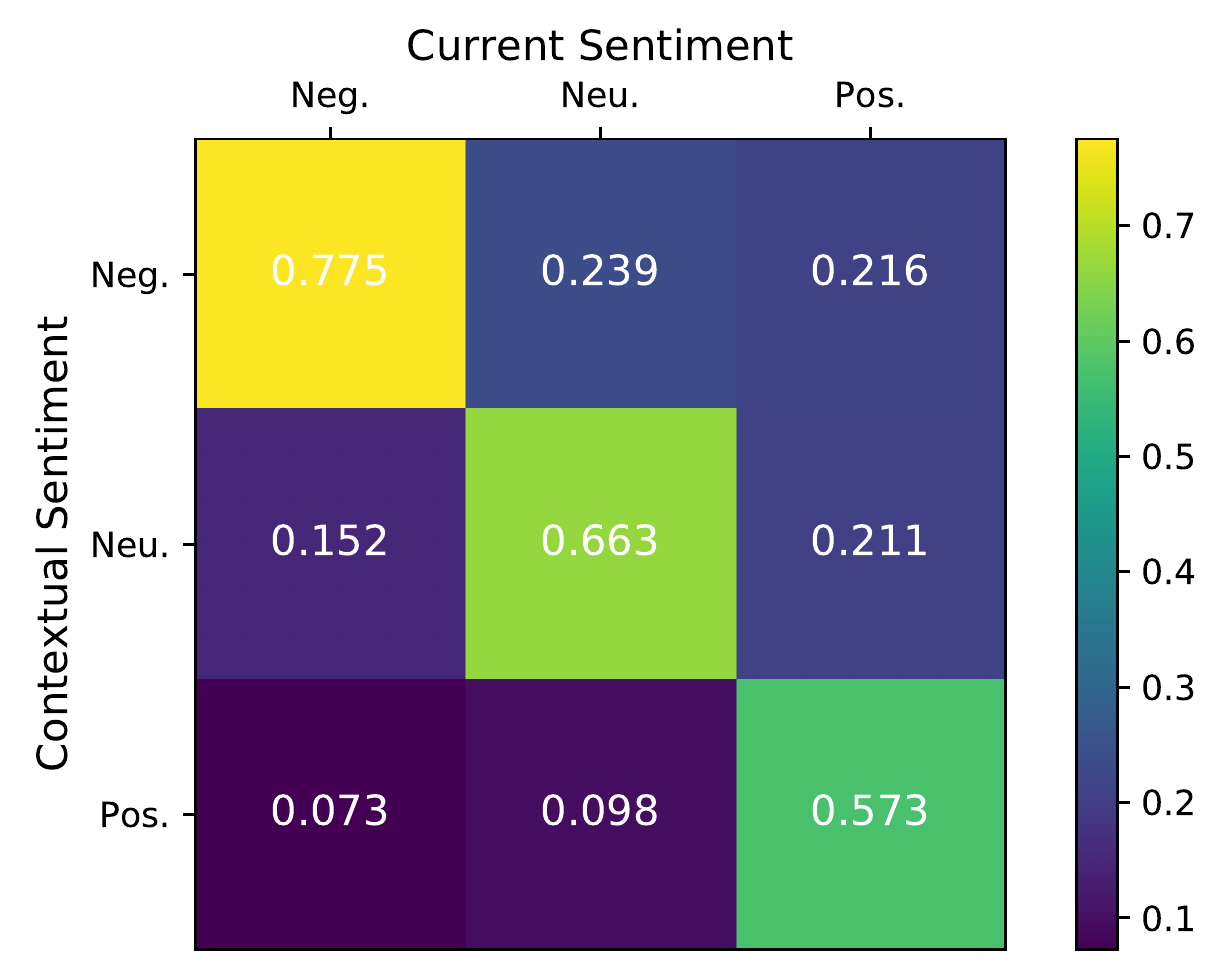}
  \caption{Probability distribution of emotion transition from context (maximum of history utterances is 5) to current utterance in CPED (Neg.: Negative, Neu.: Neutral, Pos.: Positive).}
  \label{emotion_transition}
\end{figure}

\paragraph{Experimental Results}The results of ERC in CPED is shown in Table \ref{sentiment_level_recognition}. Considering only the current utterance for emotion recognition, FastText achieves the state-of-the-art performance for \textit{negative} emotion while poor performance on the emotion classes \textit{neutral}(24.76) and \textit{positive}(0.95). Other utterance-level models have the same defects, mainly because the ability of these models to deal with \textit{label imbalance} (see Figure~\ref{Statistics}(c)) is weak. The dialogue-level models can better handle the adverse effects caused by \textit{label imbalance}. For example, \textbf{BERT+AVG+MLP} achieves the state-of-the-art performances in average accuracy and Macro-F1, since it has relatively good performance in three emotional polarities. The \textit{emotion consistence} and \textit{emotion mutation} also affect the performance of dialogue-level models. The probability of emotion transition is shown in Figure \ref{emotion_transition}. The probability of \textit{emotion mutation} in \textit{negative}, \textit{neutral} and \textit{positive} are 0.225, 0.337 and 0.427 respectively, which makes the ERC task significantly different from other long text emotion recognition tasks. In the future, it is necessary to further study the influence and challenge of \textit{emotion consistence} and \textit{emotion mutation} in dialogue to ERC task.

%(3.9$\uparrow$ compared with the state-of-the-art ERC model with current utterance and dialogue history) while BERT achieves the state-of-the-art performance for neutral sentiment (1.53$\uparrow$ compared with the state-of-the-art ERC model with current utterance and dialogue history). However, the performance of ERC models with current utterance for positive sentiment recognition is very poor (e.g. FastText achieve poor performance on the sentiment classes \textit{neutral} and \textit{positive}), mainly because the ability of these models to deal with \textit{label imbalance} (see Figure~\ref{Statistics}(c)) is weak. \textbf{BERT+AVG+MLP} achieves the state-of-the-art performances in average accuracy and Macro-F1, since it has relatively good performance in three emotional polarities. The state-of-the-art dialogue-level methods
%In addition, Chinese ERC task may be different with English ERC task, since the expression of positive emotion in Chinese may be more obscure~\cite{Bao_emo, Xiang2010Emotional}.

\section{Personalized and Emotional Conversation}
\begin{figure*}[ht]
  \centering
  \includegraphics[width=\textwidth]{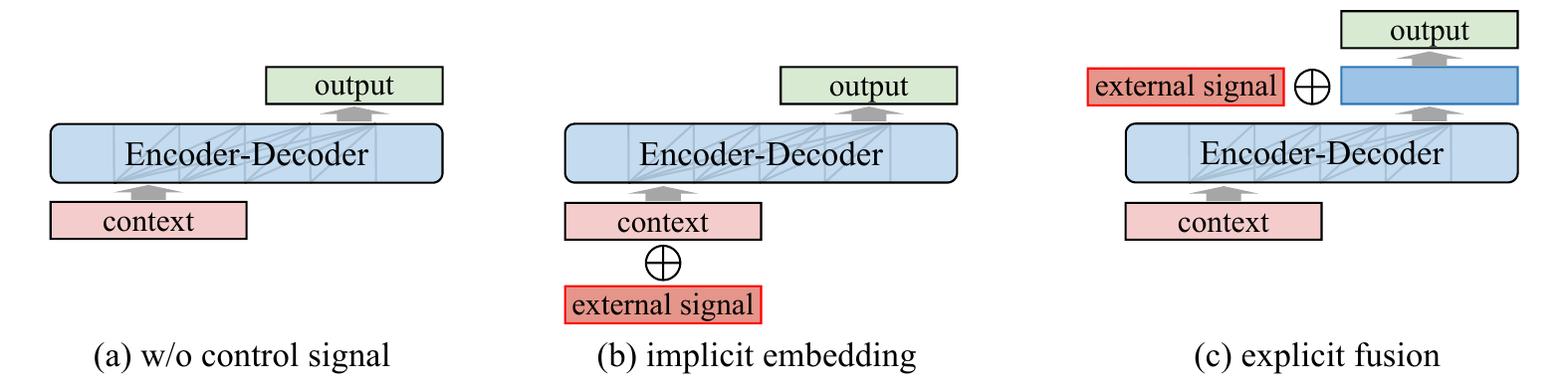}
  \caption{The generic framework of PEC. Three type of generative dialogue generation model are devised. \textit{External signal} represents emotion, personality, DA and other prior knowledge that is used to control the conversation generation.}
  \label{three_type_model}
\end{figure*}

% 个性情感对话
In this section, we provide several benchmarks for the \textbf{P}ersonalized and \textbf{E}motional \textbf{C}onversation (\textbf{PEC}) task on the proposed CPED. Conversation generation models can usually be divided into \textbf{retrieval-based} \cite{Yan2016Retrieval, Gu2020BERTRetrieval} and \textbf{generative} \cite{sordoni-etal-2015-neural, zhang2019dialogpt, Zheng2020Persona}.
As shown in Figure \ref{three_type_model}, generative conversation models can be divided into three types: (1) w/o control signal \cite{luo-etal-2018-auto, zhang2019dialogpt}, (2) implicit embedding \cite{Zheng2020Persona, EmpTransfo, zheng-etal-2021-comae}, and (3) explicit fusion \cite{zhou2018emotional, HGNN}. Generally, the latter two architectures are used for personalized conversation generation or emotional conversation generation.

\subsection{Task Definition}

We research enabling the conversation generation system to generate more anthropomorphic reply content by infusing emotion and personality at the same time.
\textbf{P}ersonalized and \textbf{E}motional \textbf{C}onversation (\textbf{PEC}) is defined as follows: Given the personalized information ($P_{R1}$ and $P_{R2}$) of two speakers, their conversation context $C$, the emotion $E_K$ and DA $D_K$ of the response to be generated, and the personalized information $P_{K}$ of the responder, the goal is to generate an anthropomorphic response $Y$.
\begin{equation}
Y = \argmax_{Y'}P(Y'|C, E_K, D_K, P_K) \label{task_definition}
\end{equation}

Particularly, context $C=\{(U_1,E_1,D_1,P_1),\cdots,(U_{K-1},$ $E_{K-1},D_{K-1},P_{K-1})\}$ contains multi-turn conversation content (i.e., utterance $U_i$), emotion $E_i$ of the associated utterance, DA $D_i$ of the associated utterance, and personalized information $P_i$ of the associated speaker.

\subsection{Baseline Models}
As shown in Figure \ref{three_type_model}, we compare several categories of generative models and our method in CPED:
\paragraph{w/o Control Signal} (1) \textbf{Seq2Seq}\cite{Ilya2014seq2seq}, the classical dialogue generation model we selected, is widely used in conversation generation. (2) \textbf{Transformer}\cite{Vaswani2017transformer}, the second model that we evaluate, is an encoder-decoder framework based on a self-attention mechanism. The transformer has been widely applied in machine translation\cite{Vaswani2017transformer}, language modeling\cite{devlin2019bert}, dialogue generation, etc.
(3) \textbf{GPT}\cite{zhang2019dialogpt} has recently gradually been used in the field of dialog generation\cite{zhang2019dialogpt, wang2020chinese}. Following \cite{wang2020chinese}, we fine-tune CDial-GPT on the CPED dataset.

\paragraph{Implicit Embedding} \textbf{\{emo+da\}-GPT} is the proposed method inspired by \cite{Zheng2020Persona} that adds word embeddings $E_{w}$, segmentation embeddings $E_{seq}$, position embeddings $E_{pos}$, emotion embeddings $E_{emo}$ and DA embeddings $E_{da}$ together as the input embeddings for GPT:
\begin{equation}
E = E_{w} + E_{emo} + E_{da}+E_{pos}+E_{seq} \label{emb_add}
\end{equation}

\paragraph{Explicit Fusion}
\textbf{GPT-\{per+emo+da\}} is the proposed method that infuses emotion $E_K$ and DA $D_K$ of the response to be generated and the personalized information $P_{K}$ of the responder. For the emotion and DA, we constructed the embedding matrix separately to obtain emotion embedding $E_g$ and DA embedding $D_g$, respectively. The embedding of personalized information is computed by a two-layer $MLP(*)$ to project $P_{K}$ to word embedding space $P_g$ as follows:
\begin{equation}
P_g = MLP(P_{K})\label{pgfnn}
\end{equation}

Subsequently, emotion embedding $E_g$, DA embedding $D_g$ and personalization embedding $P_g$ are concatenated together and then infused by a $MLP(*)$ to generate control vector $C_g$:
\begin{equation}
C_g = MLP([E_g;D_g;P_g])\label{cgfnn}
\end{equation}

We design a conditional layer to control the text generation:
\begin{equation}
O^c = O + g\odot{C_g} + (1-g)\odot{R_g} \label{o_c}
\end{equation}

where $O$ is the output of the last hidden layer of the language model (transformer or GPT, etc.). $R_g$ denotes the role of the responder, which is the word embedding of ``[speaker1]'' or ``[speaker2]''.$\odot$ is element-wise multiplication. $g\in[0,1]$ denotes the condition weight as follows:

\begin{equation}
g = \sigma{(MLP([O;C_g;R_g])}
\end{equation}

where $\sigma(*)$ is an activation function (e.g., $Tanh(*)$).

\subsection{Implementation Details}
We use transformers\footnote{\url{https://github.com/huggingface/transformers}} \cite{wolf2020transformers} and CDial-GPT \footnote{\url{https://github.com/thu-coai/CDial-GPT}} to implement the baseline model. Emotion and DA labels are added to the dictionary as special characters through the function \textit{add\_special\_tokens} of transformers for \{emo+da\}-GPT. The dimension of the word embeddings is set to 768, and the input length is $\leq$ 512 tokens. The dropout rate is set to 0.1, and the total number of training epochs is set to 120. We used the \textit{AdamW} optimizer with $\beta_{1}=0.9$, $\beta_{2}=0.999$ and the \textit{Noam} learning rate scheduler \cite{Vaswani2017transformer} with $warmup\_steps = 10000$.
We conduct experiments on Ubuntu 18.04 with 2 GeForce RTX 2080ti GPUs. 
%The number of parameters in the models used and GPU hours are shown in Table \ref{model_params_gpus}.

\iffalse
% ./CPED/runs/Jun15_23-45-39_gpu144_seq2seq
% seq2seq: 66,769,184
% ./CPED/runs/Jun15_15-50-35_gpu144_transformer
% transformer: 218,592,032
\begin{table}[htbp]
\caption{\label{model_params_gpus} Parameters and GPU hours of the models.}
\centering
\begin{tabular}{clcc}
\toprule
% Utterance_ID: 29_066_000
\textbf{Type} & \textbf{Model} & \textbf{Param.} & \textbf{GPU hours}\\
\midrule
\shortstack[c]{\small{w/o control} \\\small{signal}} & \shortstack[c]{\small{GPT}} &  \shortstack[c]{\small{95.500M}} & \shortstack[c]{\small{10h56m}}\\
\midrule
\multirow{3}{*}{\shortstack{\small{implicitly}\\\small{embedding}}} & \small{\{emo+da\}-GPT} & \small{95.525M} & \small{11h25m} \\
& \small{w/o emo} & \small{95.515M} & \small{11h16m} \\
& \small{w/o da} & \small{95.510M} & \small{11h31m} \\
\midrule
\multirow{5}{*}{\shortstack{\small{explicitly}\\\small{fusion}}} & \small{GPT-\{emo\}} & \small{97.281M} & \small{11h21m} \\
& \small{GPT-\{per\}} & \small{97.309M} & \small{11h23m} \\
& \small{GPT-\{da\}} & \small{97.286M} & \small{11h2m} \\
& \small{GPT-\{per+emo\}} & \small{97.320M} & \small{11h27m} \\
& \small{GPT-\{per+emo+da\}} & \small{99.104M} & \small{11h36m} \\
\bottomrule
\end{tabular}
\end{table}
\fi

\begin{table*}[htbp]
\caption{\label{automatic_evaluation_results} Evaluation results in CPED. The automatic evaluation includes the perplexity (\textbf{PPL}), \textbf{BLEU}, distinct-n (\textbf{D-1}, \textbf{D-2}), greedy matching (\textbf{Gre.}), embedding average (\textbf{Avg.}) and BERTscore (\textbf{BERT.}). The manual evaluation includes the content consistency (\textbf{Con.}), emotion correlation (\textbf{Emo.}) and personification capabilities (\textbf{Per.}).}
\centering
\begin{tabular}{clm{0.75cm}m{0.75cm}m{0.75cm}m{0.75cm}m{0.75cm}m{0.75cm}m{0.85cm}m{0.75cm}m{0.75cm}m{0.75cm}}
%\hline
\toprule
\multirow{2}{*}{Type} & \multirow{2}{*}{Method} & \multicolumn{7}{c}{Automatic.} & \multicolumn{3}{c}{Manual.}
    \\\cmidrule(lr){3-9}\cmidrule(lr){10-12}
           &  & \small{PPL} & \small{BLEU} & \small{D-1} & \small{D-2} & \small{Gre.} & \small{Avg.} & \small{BERT.} & \small{Con.} & \small{Emo.} & \small{Per.}\\\midrule
%& \multirow{2}{*}{\textbf{Type}} & \multirow{2}{*}{\textbf{Methods}} & \multicolumn{7}{c|}{\textbf{Automatic.}} &
%\multicolumn{3}{c}{\textbf{Manual.}}\\
%\cline{3-12}
% & & \small{PPL} & \small{BLEU} & \small{D-1} & \small{D-2} & \small{Gre.} & \small{Avg.} & \small{BERT.} & \small{Con.} & \small{Emo.} & \small{Per.}\\
%\hline

\multirow{3}{*}{\shortstack{\small{w/o control} \\\small{signal}}} & \small{Seq2seq} & \small{107.3} & \small{0.0077} & \small{0.0252} & \small{0.1846} & \small{0.4529} & \small{0.5074} & \small{0.5196} & \small{0.823} & \small{0.726} & \small{0.684}\\
& \small{Transformer} & \small{62.82} & \small{\textbf{0.1680}} & \small{0.0264} & \small{0.2031} & \small{0.4674} & \small{0.5190} & \small{0.5519} & \small{1.015} & \small{0.873} & \small{0.706}\\
& \small{GPT} & \small{20.07} & \small{0.1171} & \small{0.0482} & \small{0.2738} & \small{0.4922} & \small{0.5509} & \small{0.5629} & \small{1.118} & \small{0.963} & \small{0.760} \\
%\hline
\midrule
\multirow{3}{*}{\shortstack{\small{implicit}\\\small{embedding}}} & \small{\{emo+da\}-GPT} & \small{21.60} & \small{0.1304} & \small{0.0476} & \small{0.2785} & \small{0.4962} & \small{0.5552} & \small{0.5674} & \small{1.193} & \small{1.068} & \small{0.893} \\
& \quad \small{w/o emo} & \small{22.84} & \small{0.1252} & \small{0.0451} & \small{0.2746} & \small{0.4964} & \small{0.5564} & \small{0.5666} & \small{1.050} & \small{0.977} & \small{0.793} \\
& \quad \small{w/o da} & \small{22.09} & \small{0.1272} & \small{0.0473} & \small{0.2790} & \small{0.4962} & \small{0.5556} & \small{0.5669} & \small{1.093} & \small{0.971} & \small{0.782} \\

%\hline
\midrule
\multirow{5}{*}{\shortstack{\small{explicit}\\\small{fusion}}} & \small{GPT-\{emo\}} & \small{\textbf{17.48}} & \small{0.1342} & \small{\textbf{0.0614}} & \small{\textbf{0.3430}} & \small{0.4996} & \small{0.5588} & \small{0.5709} & \small{1.295} & \small{1.195} & \small{0.940} \\
& \small{GPT-\{per\}} & \small{18.08} & \small{0.1372} & \small{0.0592} & \small{0.3363} & \small{0.5009} & \small{0.5606} & \small{0.5715} & \small{1.308} & \small{1.042} & \small{1.043} \\
& \small{GPT-\{da\}} & \small{17.72} & \small{0.1325} & \small{0.0605} & \small{0.3389} & \small{0.5017} & \small{0.5610} & \small{0.5703} & \small{1.285} & \small{1.047} & \small{1.003} \\
& \small{GPT-\{per+emo\}} & \small{17.70} & \small{0.1403} & \small{0.0602} & \small{0.3388} & \small{\textbf{0.5026}} & \small{\textbf{0.5617}} & \small{0.5719} & \small{1.307} & \small{\textbf{1.298}} & \small{1.075} \\
& \small{GPT-\{per+emo+da\}} & \small{17.80} & \small{0.1382} & \small{0.0601} & \small{0.3404} & \small{0.5012} & \small{0.5608} & \small{\textbf{0.5722}} & \small{\textbf{1.390}} & \small{1.232} & \small{\textbf{1.237}} \\
\bottomrule
\end{tabular}
\end{table*}

\subsection{Automatic Evaluation}
\paragraph{Metrics} The perplexity (\textbf{PPL}) and \textbf{BLEU} \cite{papineni-etal-2002-bleu} are used to evaluate the relevance and fluency of the generated responses, respectively. Then, distinct-n (\textbf{D-1}, \textbf{D-2}) \cite{li-etal-2016-diversity} is applied to evaluate the degree of diversity. Greedy matching (\textbf{Gre.}), embedding average (\textbf{Avg.}) \cite{liu2016evaluate} and $F_{BERT}$ of BERTscore (\textbf{BERT.}) \cite{bertscore} are used to evaluate the semantic-level relevance of the generated responses and the reference responses.

\paragraph{Results} The results in Table \ref{automatic_evaluation_results} show that it is better to explicitly infuse the emotions and personalities of the response to be generated into the conversation model than implicitly embed them. Compared to the baseline model GPT, GPT-{emo} achieves the best PPL (2.59$\downarrow$), D-1 (0.0132$\uparrow$) and D-2 (0.0692$\uparrow$); GPT-\{per+emo\} achieves the best Gre. (0.0104$\uparrow$) and Avg. (0.0108$\uparrow$); and GPT-\{per+emo+da\} achieves the best BERT. (0.0093$\uparrow$). The results demonstrate the superiority and effectiveness of explicitly infusing emotions and personalities into open-domain conversation generation.

\subsection{Manual Evaluation}
% 人工评估：
% 回复内容维度评分：内容、话题、逻辑
% 0：
% 1：
% 2：
% 回复情感维度评分：情感、共情、
% 0：
% 1：
% 2：
% 回复个性维度评分：个性、是否像人在说话、是否像模拟的人物在说话
% 0：
% 1：
% 2：
\paragraph{Metrics} Three individual experts majoring in \textit{Chinese language and literature} were asked to evaluate the generated responses in terms of content consistency (\textbf{Con.}), emotion correlation (\textbf{Emo.}) and personification capabilities (\textbf{Per.}). \textbf{Con.} denotes the consistency of the topic and content according to the conversation context. \textbf{Emo.} denotes the emotional relevance and rationality of the response generated by the dialogue system. \textbf{Per.} denotes the personification capabilities of the dialogue system and is applied to measure the human-like expression ability. The rating scale is $(0, 1, 2)$, where 0 means the worst and 2 means the best.
%内容维度的kappa分： 0.6581041755990438
%情感维度的kappa分： 0.6318795909954932
%个性维度的kappa分： 0.6455342762270956
\paragraph{Results} Two hundred dialogues were randomly sampled from the test set of CPED for manual evaluation. Fleiss' kappa\cite{fleiss1971measuring} is calculated to measure the inter-rater consistency for \textbf{Con.}, \textbf{Emo.}, and \textbf{Per.}, which are 0.658, 0.632 and 0.646, indicating substantial annotation agreement respectively. 
%All the evaluation metrics have ``moderate agreement'' or ``substantial agreement''.
Table \ref{automatic_evaluation_results} shows the results of the manual evaluation in terms of content, emotion and personification. We observe that GPT-\{per+emo+da\} achieves the best \textbf{Con.} (0.272$\uparrow$) and the best \textbf{Per.} (0.477$\uparrow$) compared with GPT while GPT-\{per+emo\} achieves the best \textbf{Emo.} (0.335$\uparrow$). This demonstrates that ``explicit fusion'' can effectively benefit the conversation generation model to generate more anthropomorphic responses. Furthermore, explicitly specifying the emotion and personality of the responses will improve the emotional expression ability and personality expression ability of the dialogue system.

\subsection{Case Study}
In Table \ref{sample_responses}, we present an example of the answers generated by the baseline models to give insight into whether the emotion and personality of the generated responses are expressed appropriately. The table shows that \textbf{GPT-\{per+emo+da\}} can generate highly anthropomorphic responses (e.g., \begin{CJK}{UTF8}{gbsn}你想得美。\end{CJK}(When pigs fly!)) with appropriate emotion and personality while the \textbf{GPT} could not express the emotion ``\textit{anger}'' with the generated response ``\begin{CJK}{UTF8}{gbsn}我还是想要你的。\end{CJK}(I still want yours.)''. 
In other words, when the emotion and DA of a response are generated and the personalities of the responder are explicitly infused into the conversation generation model, the model can perform with a high personification level and suitable emotional expression.

\begin{table*}[htbp]
\caption{\label{sample_responses} Sample responses generated by the baseline models. sd: statement-non-opinion, c: command, dag: disagreement, q: question, sv: statement-opinion.} %  The personality of the responder is (male, high, high, high, high, low) in terms of (Gender, Neuroticism, Extraversion, Openness, Agreeableness, Conscientiousness).
\centering
\begin{tabular}{c|l|m{1.5cm}|m{0.5cm}|m{9.9cm}}
\toprule
% Utterance_ID: 29_066_000
 & \textbf{Speaker} & \textbf{Emotion} & \textbf{DA} & \textbf{Utterance} \\
\midrule

\multirow{11}{*}{\shortstack{\small{Dialogue} \\\small{History}}} & \small{Zeng Xiaoxian} & \small{neutral} & \small{sd} & \begin{CJK}{UTF8}{gbsn}\small{要挽留美嘉,}\end{CJK}\small{(To retain Meijia,)}\\
& \small{Zeng Xiaoxian} & \small{positive(o)} & \small{sd} & \begin{CJK}{UTF8}{gbsn}\small{当然是用子乔的照片了。}\end{CJK}\small{(Certainly, I used Ziqiao's photo.)}\\
& \small{Hu Yifei} & \small{anger} & \small{sd} & \begin{CJK}{UTF8}{gbsn}\small{你盗用我的创意。}\end{CJK}\small{(You theft of my idea.)}\\
& \small{Zeng Xiaoxian} & \small{neutral} & \small{sd} & \begin{CJK}{UTF8}{gbsn}\small{奇怪了。}\end{CJK}\small{(It's weird.)}\\
& \small{Zeng Xiaoxian} & \small{neutral} & \small{sd} & \begin{CJK}{UTF8}{gbsn}\small{我们是在比谁更了解美嘉的心思。}\end{CJK}\small{(We are comparing who knows the mind of Meijia better.)}\\
& \small{Zeng Xiaoxian} & \small{neutral} & \small{sd} & \begin{CJK}{UTF8}{gbsn}\small{又不是比创意。}\end{CJK}\small{(Not comparing the creativity.)}\\
& \small{Zeng Xiaoxian} & \small{relaxed} & \small{sd} & \begin{CJK}{UTF8}{gbsn}\small{要放当然放我的。}\end{CJK}\small{(Of course, place mine.)}\\
& \small{Hu Yifei} & \small{negative(o)} & \small{c} & \begin{CJK}{UTF8}{gbsn}\small{把我这张也放进去。}\end{CJK}\small{(Place mine into there too.)}\\
& \small{Zeng Xiaoxian} & \small{negative(o)} & \small{dag} & \begin{CJK}{UTF8}{gbsn}\small{不行。}\end{CJK}\small{(No way.)}\\
& \small{Zeng Xiaoxian} & \small{depress} & \small{q} & \begin{CJK}{UTF8}{gbsn}\small{那么有效的话算谁的?}\end{CJK}\small{(If it is effective, Whose credit is this?)}\\
& \small{Hu Yifei} & \small{anger} & \small{sv} & \begin{CJK}{UTF8}{gbsn}\small{当然算我的，是我的主意。}\end{CJK}\small{(Of course it counts as mine, this is my idea.)}\\
\midrule
\small{Golden} & \small{Zeng Xiaoxian} & \small{anger} & \small{dag} & \begin{CJK}{UTF8}{gbsn}\small{没门！}\end{CJK}\small{(No way!)}\\
\midrule
\shortstack{\small{w/o control} \\\small{signal}} & \small{GPT} & \small{-} & \small{-} & \begin{CJK}{UTF8}{gbsn}\small{我还是想要你的。}\end{CJK}\small{(I still want yours.)}\\
\midrule
\multirow{3}{*}{\shortstack{\small{implicitly}\\\small{embedding}}} & \small{\{emo+da\}-GPT} & \small{-} & \small{-} & \begin{CJK}{UTF8}{gbsn}\small{我就不用你的。}\end{CJK}\small{(I will not use yours.)}\\
& \small{w/o emo} & \small{-} & \small{-} & \begin{CJK}{UTF8}{gbsn}\small{你是我的。}\end{CJK}\small{(You belong to me.)}\\
& \small{w/o da} & \small{-} & \small{-} & \begin{CJK}{UTF8}{gbsn}\small{你在哪？}\end{CJK}\small{(Where are you?)}\\
\midrule
\multirow{5}{*}{\shortstack{\small{explicitly}\\\small{fusion}}} & \small{GPT-\{emo\}} & \small{anger} & \small{-} & \begin{CJK}{UTF8}{gbsn}\small{我决定了。}\end{CJK}\small{(I've made a decision.)}\\
& \small{GPT-\{per\}} & \small{-} & \small{-} & \begin{CJK}{UTF8}{gbsn}\small{怎么算？}\end{CJK}\small{(How to judge?)}\\
& \small{GPT-\{da\}} & \small{-} & \small{dag} & \begin{CJK}{UTF8}{gbsn}\small{行吗？}\end{CJK}\small{(Will that work?)}\\
& \small{GPT-\{per+emo\}} & \small{anger} & \small{-} & \begin{CJK}{UTF8}{gbsn}\small{我一定会好好考虑。}\end{CJK}\small{(I will definitely think about it.)}\\
& \small{GPT-\{per+emo+da\}} & \small{anger} & \small{dag} & \begin{CJK}{UTF8}{gbsn}\small{你想得美。}\end{CJK}\small{(When pigs fly!)}\\
\bottomrule

\end{tabular}
\end{table*}

\section{Applications and Limitation of CPED}
\subsection{Applications}
CPED allows evaluation of both conversational cognitive tasks and conversation generation tasks, e.g. speaker modeling, personality recognition in conversations, emotion recognition in conversations, DA recognition in conversations, emotion prediction for response, emotional conversation generation, personalized conversation generation, empathetic conversation etc. By being multimodal, CPED can also be applied in multimodal personality or emotion recognition, multimodal conversation generation. It will play a positive role in promoting the development of cognitive intelligence.

% 应用与局限性
\subsection{Ethical Considerations}
\paragraph{Data and Privacy}
All the dialogue materials are based on TV dramas (publicly available source: Tencent Video\footnote{\url{https://v.qq.com}}, Youku Video\footnote{\url{https://youku.com}}, iQiyi Video\footnote{\url{https://iqiyi.com}}) in which the names of the characters are all fictitious. Correspondingly, the personalities are also marked from the performance of the characters in the TV dramas. The video and audio clips are licensed under Copyright Law of the People's Republic of China. According to the privacy issues, copyright claims, terms of service of Tencent Video, Youku Video and iQiyi Video, we only release the textual dataset with audio features and video features.

\begin{table}[htbp]
\caption{\label{statistics_risk_responses} Statistics of the negative responses and dangerous responses generated by the baseline models. \textbf{Neg.} is the proportion of negative responses, and \textbf{Dan.} is the proportion of angry responses.}
\centering
\begin{tabular}{m{1.7cm}lcc}
\toprule
% Utterance_ID: 29_066_000
\textbf{Type} & \textbf{Model} & \textbf{Neg.} & \textbf{Dan.}\\
\midrule
\small{w/o control signal} & \small{GPT} &  \small{1.0\%} & \small{0.5\%}\\
\midrule
\multirow{3}{*}{\shortstack{\small{implicitly}\\\small{embedding}}} & \small{\{emo+da\}-GPT} & \small{3.5\%} & \small{0.0\%} \\
& \small{w/o emo} & \small{1.5\%} & \small{0.0\%} \\
& \small{w/o da} & \small{3.0\%} & \small{0.5\%} \\
\midrule
\multirow{5}{*}{\shortstack{\small{explicitly}\\\small{fusion}}} & \small{GPT-\{emo\}} & \small{4.5\%} & \small{0.5\%} \\
& \small{GPT-\{per\}} & \small{3.5\%} & \small{0.5\%} \\
& \small{GPT-\{da\}} & \small{0.5\%} & \small{0.0\%} \\
& \small{GPT-\{per+emo\}} & \small{3.5\%} & \small{1.0\%} \\
& \small{GPT-\{per+emo+da\}} & \small{2.5\%} & \small{1.5\%} \\
\bottomrule

\end{tabular}
\end{table}

% 讨论电视剧数据与自然对话数据的差异
\paragraph{Difference between television conversation and natural conversation}
Similar to FriendsPersona\cite{jiang2020friendspersona} and MELD~\cite{poria2019meld}, CPED is derived from TV shows. According to the book \textit{Television Dialogue: The sitcom Friends vs natural conversation}~\cite{Quaglio2009Television}, television conversations and natural conversations are basically the same in terms of linguistic features. However, television conversations tend to present a limited set of scenarios, interaction types and topic categories. To this end, when we select TV series, we try to cover different scenes of daily life as much as possible (see Table \ref{annotation_label_table}). In addition, due to the entertainment characteristics of TV shows, screenwriters often use expletive, slang, appellation and other language means to achieve humorous effects, in order to make the language in TV shows more authentic and attractive. Therefore, the emotion distribution of television conversations may be different from natural conversations (see Figure \ref{Statistics}-(c)). When using this dataset, you can consider joint pre-training with unlabeled natural conversation datasets to alleviate this problem.

\paragraph{Potential bias and Ethical Risk}
We realize that if the model learns anthropomorphic expression ability, it may also learn the negative expressions or dangerous expressions brought about by personality.
\textit{Negative responses} represent those responses that make the emotions of both sides of the conversation develop in a worse direction.
\textit{Dangerous responses} represent those types of responses that involve suicide, abetting others to commit suicide, intimidation, etc.
As shown in Table \ref{statistics_risk_responses}, we randomly selected 200 samples from the test set and counted the proportions of \textit{negative responses} and \textit{dangerous responses}. It is foreseeable that by improving the personification level of the dialogue generation model, it is also possible for the dialogue model to learn those risk responses. When using the CPED dataset, users should consider how to reduce the possibility of risk responses from the dialogue system while improving the level of personification of the dialogue system.

\section{Conclusion and Future Work}
% 分点描述Future Work，把数据集的未来价值阐述得更加清晰
In this paper, we proposed a challenging dataset CPED for conversational AI, a large-scale Chinese personalized and emotional dialogue dataset containing more than 11K dialogues with 392 speakers from 40 TV shows. CPED contains abundant prior information about emotions, personalities, dialog acts and other items. We introduce three challenging tasks for conversational AI research in CPED, including personality recognition, emotion recognition in conversations as well as personalized and emotional conversation generation. The evaluation results of the baseline models are initial but indicative. In Chinese conversations, the tasks of personality recognition and emotion recognition need to be further with linguistic characteristics and psychological knowledge, e.g. differences in different personality dimensions, differences of the demonstrated personalities or emotions in conversations between speakers and different characters, linguistic characteristics of different sentiment polarities and etc. For personalized and emotional conversation, explicitly infusing emotions, personalities and dialog acts of the response to be generated can improve the personification level and emotional expression of a dialogue system. We believe that CPED can help researchers study both the cognitive processing in conversations and the personalized and emotional conversation (PEC) task.
Based on the abundant emotions, personalities, and multimodal contexts of CPED, future work can explore the following: (i) modeling or recognition of speakers' personality and emotion, (ii) prediction of responded emotion and personality, (iii) personalized and emotional conversation generation using multimodal contexts, (iv) pretrained PEC model for empathetic conversation or mental health support, etc.
\iffalse
\begin{itemize}
\item Modeling or recognition of speakers' personality and emotion.
\item Prediction of responded emotion and personality.
\item Personalized and emotional conversation generation using multimodal contexts.
\item Pretrained PEC model for empathetic conversation or mental health support, etc.
\end{itemize}
\fi

\bibliography{custom}
\bibliographystyle{IEEEtran}

%\appendix % for AAAI or ACL, NAACL, EMNLP
% \renewcommand{\appendixname}{Appendix~\Alph{section}}
\begin{appendices} % for IEEE Trans Template
% 附录额外讨论
\section{Relationships between Emotions and DAs}

\begin{figure}[ht]
  \centering
  \includegraphics[width=0.48\textwidth]{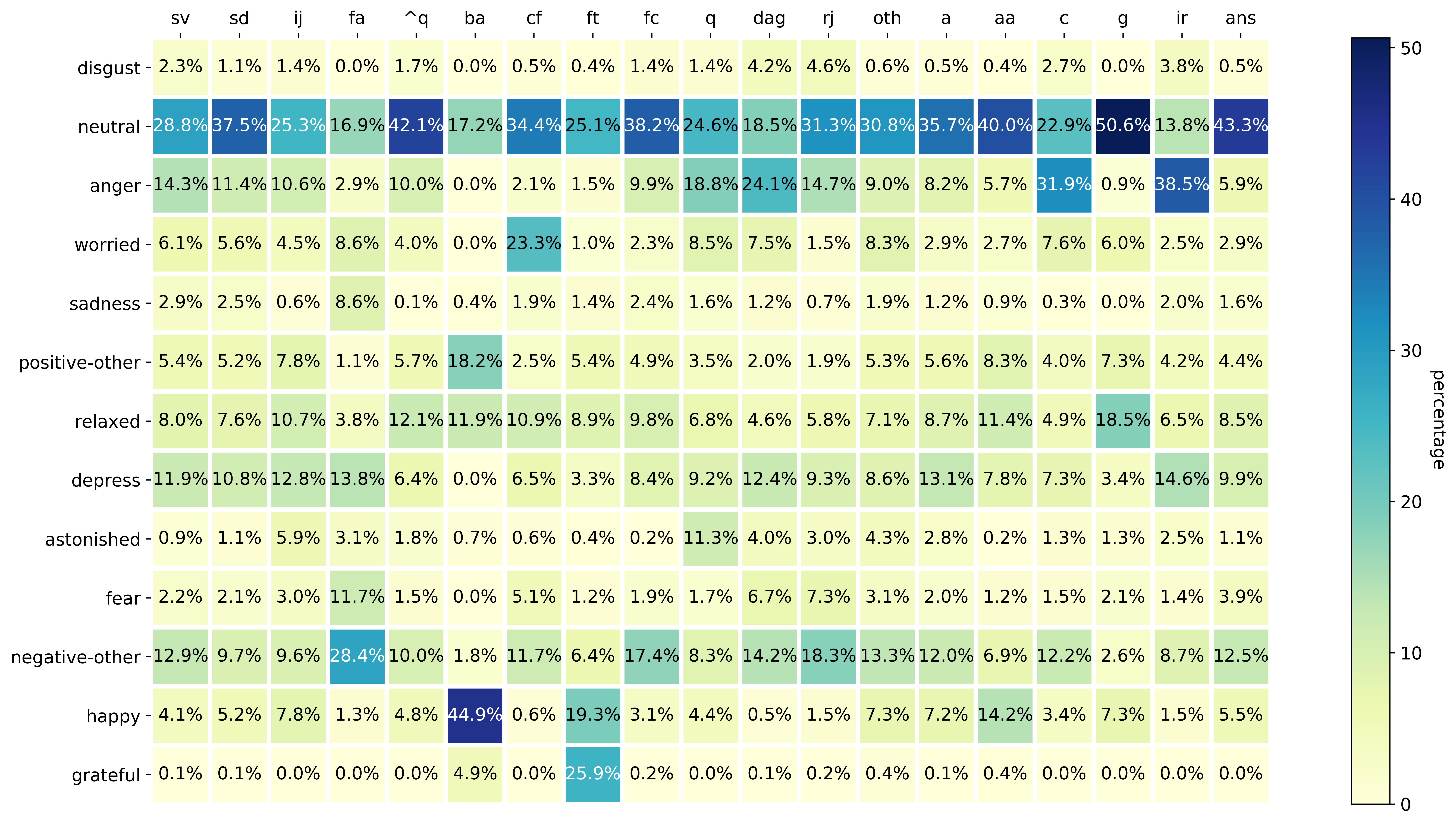}
  \caption{Relation between the Emotions and DAs.}
  \label{eda_relation}
\end{figure}

Furthermore, the relationships between emotions and DAs are shown in Figure \ref{eda_relation}. According to the statistics, most DAs will appear at the same time as ``\textit{neutral}''. ``\textit{Appreciation (ba)}'' is mainly related to ``\textit{happy}'' (44.9\%). ``\textit{Thanking (ft)}'' has an obvious correlation with ``\textit{happy}'' and ``\textit{grateful}''. ``\textit{Disagreement (dag)}'', ``\textit{command (c)}'' and ``\textit{irony (ir)}'' have significant correlations with ``\textit{angry}''. ``\textit{Comfort (cf)}'' has an obvious correlation with ``\textit{worried}''.

\end{appendices}

\end{document}